\documentclass[manuscript,screen,nonacm]{acmart}

\usepackage{amsmath}
\usepackage{booktabs}
\usepackage{multirow}
\usepackage{array}
\usepackage{makecell}
\usepackage{enumitem}
\usepackage{adjustbox}
\usepackage{tabularx}
\usepackage{longtable}
\usepackage{algorithm}
\usepackage{algpseudocode}
\usepackage[normalem]{ulem}
\usepackage{placeins}

\makeatletter
\newcommand{\abstractnote}[2][1]{%
  \begingroup
  \renewcommand{\thefootnote}{\@fnsymbol\c@footnote}%
  \footnotemark[#1]\footnotetext[#1]{#2}%
  \endgroup}
\ifdefined\AtBeginMaketitle
  \AtBeginMaketitle{\setcounter{footnote}{1}}
\fi
\ifdefined\@titlenotes
  \g@addto@macro\@titlenotes{\stepcounter{footnote}}
\fi
\makeatother

\begin{document}

\title{Large Models for Small Devices: Recent Advances and Empirical Analysis of Edge AI Deployment}

\author{Subhransu Das}
\email{das.411@osu.edu}

\author{Jiaming Cheng}
\authornote{The author contributed to this work as an external collaborator and holds no formal appointment at The Ohio State University.}

\author{Arnav Kumar}
\email{kumar.1178@osu.edu}

\author{Sadia Afrose}
\email{afrose.4@osu.edu}

\author{Mingzhe Han}
\email{han.1453@osu.edu}

\author{Michael Silagy}
\email{silagy.4@osu.edu}

\author{Shreya Palande}
\email{palande.3@osu.edu}

\author{Brijesh Soni}
\email{soni.152@osu.edu}

\author{Rajiv Ramnath}
\email{ramnath.6@osu.edu}
\affiliation{%
  \institution{The Ohio State University}
  \department{Department of Computer Science and Engineering}
  \city{Columbus}\state{OH}\country{USA}}

\renewcommand{\shortauthors}{Das et al.}

\begin{abstract}
Running large AI models on resource-constrained edge devices requires model compression to reduce model size and computation. What compresses well, however, need not deploy well. We survey dozens of recent works that report compression results on real hardware and extract practical deployment guidelines from them. Following these guidelines, we deploy compact language and image models on GPU, CPU, and Raspberry Pi platforms across question answering and image segmentation\abstractnote{Parts of this work were presented at the IEEE Consumer Communications \& Networking Conference (CCNC), held in Las Vegas, NV, USA, in January 2026~\cite{Our_CCNC_2026work}.}.

No single technique wins across tasks. For question answering, Qwen3.5 0.8B reaches 93.85 SQuAD F1 and 92 EM under \texttt{Q5\_K\_M} GGUF quantization, while structured pruning at the same precision costs 16 F1 at a 1\% ratio. For segmentation, the ranking reverses: default quantization leaves parameters and MACs unchanged, whereas pruning cuts model size by nearly 80\% at near-constant mIoU. Pruning can even inflate the deployed artifact by 21--49\% by breaking k-quant super-block alignment; combined with longer, less format-compliant outputs, this raises Raspberry Pi latency up to 3.4$\times$. Compression can also manufacture the appearance of competence rather than destroy it visibly: one LoRA-recovered variant stays fully parseable and holds 71\% strict BoolQ accuracy while sending 97 of 100 predictions to a single class, at 52.6\% balanced accuracy. We explain these effects through neural-flow graph analysis and prefill--decode-level latency decomposition, and condense them into task-specific deployment research directions. The right technique depends on the task, the model, and the hardware.

Our experiment code and artifacts are open-sourced on GitHub.\abstractnote[3]{\url{https://github.com/Arnavvvkumar/deployment}}
\end{abstract}


\received{15 June 2026}

\maketitle

 \section{Introduction} \label{sec:intro}

Whether an AI model can be deployed at the edge is decided by whether its memory, compute, and energy demands fit the target device. The models now carrying applications in healthcare, autonomous driving, and customer support \cite{zhao2025autonomousdriving, fahim2025aihealthcare, nicolescu2022chatbots} are rarely designed with those budgets in mind. For this reason, many promising models, such as DeepLabV3+, SegFormer, Mask2Former, and the Segment Anything Model (SAM), which could be highly beneficial for edge applications like post-flood aerial scene understanding on FloodNet imagery \cite{rahnemoonfar2021floodnet,chen2018deeplabv3plus,xie2021segformer,cheng2022mask2former,kirillov2023segmentanything}, arrive with no evaluation on edge hardware --- SegFormer's authors acknowledge that even for their smallest model, at 3.7M parameters, ``it is unclear whether it can work well in a chip of edge device with only 100k memory'', and leave the question to future work \cite{xie2021segformer}.

Model compression techniques, such as pruning, quantization, tensor decomposition and knowledge distillation, address this issue by reducing model size, memory usage and computational cost. Pruning identifies less important components in a network and masks them during computation (unstructured pruning) or physically removes them from the network
(structured pruning) \cite{fang2023depgraph} \cite{fang2024isomorphic}. Quantization reduces the numerical precision of weights and activations, while tensor decomposition approximates large weight tensors using smaller low-rank factors. Knowledge distillation trains a smaller student model to mimic the behavior of a larger teacher model \cite{setyanto2025} \cite{wang2024}. Together, these methods can make deep models more lightweight and suitable for resource-constrained edge devices.

However, compression does not transfer uniformly: a method that works well for one task may cause a large accuracy drop or limited speedup for another task. This motivates the need for both a deployment-oriented survey and an empirical study. The survey summarizes studies that report measured deployment results and identifies practical guidelines, while the empirical study evaluates different compression techniques across multiple downstream tasks and domains to understand their real deployment behavior.

Compressed large language models (LLMs) are now the center of edge-deployment research: local inference offers low latency and privacy without connectivity. One line of surveys organizes the edge LLM stack itself. Liang et al.~\cite{survey2_LLM_compression_on_Edge_IoTJ_2026} catalogue compression methods for language models in edge systems; Zheng et al.~\cite{Survey5_EdgeLLM_ACM2025} organize edge LLMs by design, execution, and application; Qu et al.~\cite{survey7_EI_LLM_CST2025} center mobile edge intelligence, distributed learning, and 6G-enabled deployment; and Chen et al.~\cite{survey8_very_recent_2026} treat network-edge inference through deployment architecture and resource management. Together these map the design space of compression and serving, but each compares methods as its source papers reported them rather than under one measurement protocol.

A second line widens the scope past language models. Navardi et al.~\cite{survey4_Navardi2025GenAIEdge} review generative AI on resource-constrained devices across software optimization, hardware optimization, and deployment frameworks. Wang et al.~\cite{survey6_wang2025cognitiveedgecomputingcomprehensive} frame cognitive edge computing as a route to running large models and AI agents on pervasive edge systems, and in a companion survey \cite{survey3_empoweringEdgeAI_2025ACM} cover on-device AI more broadly, including applications, technical challenges, and implementation strategies. Liu et al.~\cite{survey9_recent_CST2026} treat edge--cloud collaborative computing and distributed model optimization. This breadth costs resolution on any single deployment target: these surveys describe what can be done rather than what a given method costs on a given device.

Across both lines, three practical gaps remain, and they organize our survey and empirical study.

\begin{itemize}
    \item First, most existing surveys mainly provide theoretical summaries of prior work and do not systematically distinguish studies that report actual deployment results on resource-constrained hardware, whereas our survey focuses on recent works that apply compression techniques and report measured deployment results, in most cases including acceleration numbers on real hardware platforms.

    \item Second, existing surveys often aggregate results from different papers, so the numbers they compare were never measured under one protocol. In this work, we report our own measurements: pruning, quantization, and LoRA recovery evaluated on GPU, CPU, and Raspberry Pi across natural language processing tasks (SQuAD, BoolQ, and Natural Questions), and semantic segmentation models compressed with pruning, quantization, and knowledge distillation in our prior deployment pipeline, reported with task-relevant metrics such as mean accuracy and mIoU.

    \item Third, existing studies rarely analyze on-device LLM inference as a two-stage process consisting of prefill and decode phases. In this work, we analyze deployed inference as a compute-bound prefill and a memory-bandwidth-bound decode, and use that split to explain deployment-level behavior that end-to-end latency alone hides, including why structural pruning raised per-token decode cost in all 18 matched baseline-versus-pruned pairs and slowed every internally controlled CPU comparison, despite shrinking the stored checkpoint.

\end{itemize}

\begin{figure*}[!t]
    \centering
    \includegraphics[
        width=.8\textwidth,
        height=0.4\textheight,
        keepaspectratio
    ]{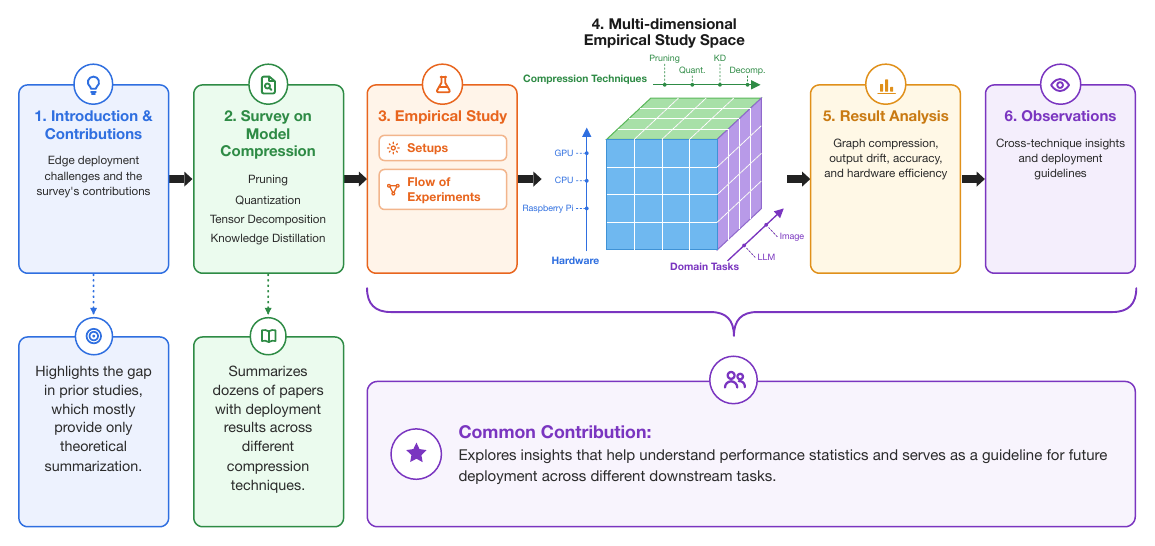}
    \caption{Organization of the paper}
    \label{fig:organization}
\end{figure*}

Fig.~\ref{fig:organization} maps the organization of this paper.

\section{Contributions}

\begin{itemize}
\item We pair a survey of more than twenty works that report measurements on real constrained hardware with a cross-task deployment study: three approximately 1B-parameter LLM families under structured pruning, GGUF quantization, and LoRA recovery on GPU, CPU, and Raspberry Pi, and six segmentation models under pruning, quantization, and knowledge distillation from our prior deployment pipeline. The pairing shows when nominal compression delivers device-level size and latency gains and when it does not.

\item We identify pruning-induced forgetfulness --- the measured loss of task quality and output discipline after structural removal: pruning as little as 1\% of MLP channels already costs substantial quality across all three model families we evaluate, and post-pruning LoRA fine-tuning recovers only part of it. To account for this behavior, we model the neural network as a flow graph and derive a first-order, layer-additive measure of the local signal distortion introduced by pruning. Computed for every pruned configuration, the measure orders damage within a model family --- rank-matching the measured SQuAD F1 loss for both Qwen trajectories --- and resolves where the damage lands: the gate projection carries $77\%$ of the distortion in Qwen3.5 while Qwen2.5 concentrates $67\%$ on the down projection, so raw magnitudes remain family-specific. This measured locality motivates the post-pruning fine-tuning we evaluate.

\item We analyze on-device LLM inference as a two-stage process consisting of prefill and decode phases, and measure the split directly from serving logs over 18 matched baseline-versus-pruned run pairs. The added latency concentrates in the generation phase in every pair: pruned models generate more tokens (up to $2.7\times$ for Gemma), and once structured pruning breaks $k$-quant block alignment and the deployable weight set grows by 21--49\%, each token also costs more --- though the per-token increase stays well below the byte growth. This stage-wise account links the compression method, the size of the deployed artifact, and the memory bandwidth of the target device to the latency measured on each platform.

\item We identify a prior-collapse failure mode on BoolQ: under compression and recovery a model can remain fully parseable and hold $71\%$ strict accuracy while directing $97$ of $100$ predictions to a single class, at $52.6\%$ balanced accuracy. The collapse direction depends on the model, so strict accuracy ranks two collapsed configurations in the opposite order to balanced accuracy. Reporting prediction skew alongside balanced accuracy exposes the discrimination that strict accuracy hides.

\end{itemize}

\section{Overview of Compression Techniques and Recent Advances}\label{sec:overview}

This section describes four major compression techniques: pruning, quantization, tensor decomposition, and knowledge distillation. We review selected works that report deployment results to understand how these techniques perform under real edge-device constraints.

\subsection{Pruning}

Pruning is a model compression technique that removes less important weights or structural components from a neural network with the aim of preserving its task performance. Let $f(x;\Theta)$ be a neural network that maps an input $x$ to a prediction using parameters $\Theta$, and let $M$ be a binary pruning mask applied to those parameters or structures. The pruning problem is commonly formulated as the constrained optimization
\begin{equation}
\min_{\Theta, M} \; \mathcal{L}\big(f(x; M \odot \Theta), y\big)
\quad \text{s.t.} \quad |M|_0 \leq K,
\label{eq:pruning_optimization}
\end{equation}
where $\odot$ is the element-wise product so that $M \odot \Theta$ is the pruned model, $\mathcal{L}$ is the task loss between the prediction and the ground-truth target $y$, $|M|_0$ counts the retained components, and $K$ sets the target model size or sparsity budget. In practice, the mask $M$ is selected using saliency scores \cite{fang2024isomorphic}, such as weight magnitude, Taylor importance, or empirical Fisher information, so that components with lower importance are removed first. This targeted removal reduces the model's parameter count and multiply--accumulate operations.

To consolidate our discussion, we survey recent work along three deployment-relevant axes: adaptive structured pruning, fine-grained unstructured pruning, and compound compression pipelines. The first group uses structured pruning to adapt models either to the local deployment environment or to the latency behavior of a target device. Ma et al. propose OCAP, which removes channels less relevant to locally observed classes and reduces latency by more than 50\% while improving retained-class accuracy by up to 20\% on NVIDIA Jetson platforms \cite{ma2023ocap}, while Belhadi et al. introduce LightPrune, which uses a differentiable latency estimator to reduce MobileNetV2 size by up to 3$\times$ and improve inference latency by 2.2$\times$ on Jetson Nano \cite{belhadi2025lightprune}. The second group focuses on fine-grained unstructured pruning, where individual weights are removed while the overall network topology is mostly preserved. Crespí-Castañer et al. use auxiliary Morphological Neural Networks to generate pruning masks for dense layers and remove up to 98\% of weights, achieving 1025.6 inferences/J on Raspberry Pi Pico, corresponding to more than a 46$\times$ improvement in energy efficiency when combined with a custom C-based inference engine \cite{crespi2025densepruning}. The third group treats pruning as one stage of a broader compression pipeline, where Chang et al. combine mixed pruning with 8-bit quantization and FPGA acceleration to achieve 30$\times$ VGG16 compression and up to 24.5$\times$ better energy efficiency on Xilinx ZCU102 \cite{chang2021mixedpruning}, Singh et al. evaluate both unstructured weight pruning and structured unit pruning, then combine pruning with INT8 TensorFlow Lite quantization to reduce model size by about 10$\times$ in software-based evaluations \cite{singh2023pruningquantization}, and Dai et al. use Bayesian inference to select suitable channel pruning ratios before applying Tucker tensor decomposition to factorize high-dimensional convolutional tensors, reducing parameters by up to 34.69$\times$ and FLOPs by 8.31$\times$ on a Samsung Edge platform with a Mali-T880 GPU \cite{dai2022compressing}. Pruning, therefore, reduces the storage, computation, and energy burden of edge models, but these benefits come with practical limits, since aggressive sparsity can degrade accuracy, and the expected latency or energy gains appear only when the target runtime or hardware can efficiently execute the pruned model. Table~\ref{tab:compression_edge_summary} summarizes the reviewed pruning methods.

\newcolumntype{L}[1]{>{\raggedright\arraybackslash}p{#1}}
\newcolumntype{C}[1]{>{\centering\arraybackslash}p{#1}}

\newcommand{\cmark}{\scalebox{0.68}{$\checkmark$}}
\newcommand{\xmark}{\scalebox{0.68}{$\times$}}

\begingroup
\scriptsize
\setlength{\tabcolsep}{4pt}
\renewcommand{\arraystretch}{1.24}

\newcolumntype{Y}{>{\raggedright\arraybackslash}p{\dimexpr(\textwidth-10\tabcolsep-2\arrayrulewidth)/5\relax}}
\begin{longtable}{|Y Y Y Y Y|}
\caption{Comparison of Model Compression Methods for Edge Deployment}
\label{tab:compression_edge_summary} \\

\hline
\textbf{\scriptsize Concept} &
\textbf{\scriptsize Models} &
\textbf{\scriptsize Edge Devices} &
\textbf{\scriptsize Downstream Task} &
\textbf{\scriptsize Performance Note}  \\
\hline\hline
\endfirsthead

\multicolumn{5}{l}{\scriptsize\itshape Table~\ref{tab:compression_edge_summary} continued from previous page} \\
\hline
\textbf{\scriptsize Concept} &
\textbf{\scriptsize Models} &
\textbf{\scriptsize Edge Devices} &
\textbf{\scriptsize Downstream Task} &
\textbf{\scriptsize Performance Note}  \\
\hline\hline
\endhead

\hline
\multicolumn{5}{r}{\scriptsize\itshape continued on next page} \\
\endfoot

\hline
\endlastfoot

\multicolumn{5}{|l|}{\textbf{\small Pruning}} \\
\hline
On-device class-aware pruning (OCAP) \cite{ma2023ocap} &
VGG-16, ResNet-56, MobileNetV2 &
Jetson Nano, TX2, AGX Xavier &
Classification &
$>50\%$ latency reduction, with accuracy improvements reported up to 20\%. \\[2mm]

Latency-aware structured pruning \cite{belhadi2025lightprune} &
MobileNetV2 &
Nvidia Jetson Nano &
Classification &
3$\times$ model-size reduction and 2.2$\times$ latency speed-up. \\[2mm]

Morphological pruning (MHP/MEP) \cite{crespi2025densepruning} &
LeNet-300-100, LeNet-5 &
Raspberry Pi Pico, ESP32-E &
Classification &
Removes up to $\sim$98\% of weights and achieves  $>$ 46$\times$ energy efficiency gain. \\[2mm]

Mixed pruning with 8-bit quantization \cite{chang2021mixedpruning} &
VGG16, FCN8s &
Xilinx ZCU102 FPGA &
Classification, segmentation &
30$\times$ compression for VGG16 and 24.5$\times$ improvement in energy efficiency. \\[2mm]

Channel pruning with Tucker decomposition \cite{dai2022compressing} &
VGG-16, AlexNet, ResNet-18 &
Samsung Edge device with Mali-T880 GPU &
Classification &
34.69$\times$ parameter reduction and 8.31$\times$ FLOPs reduction. \\[2mm]

Weight/unit pruning with INT8 quantization \cite{singh2023pruningquantization} &
Custom 4-layer MLP; standard CNNs &
TensorFlow Lite-based edge setting &
Classification &
10$\times$ model-size reduction. \\[1mm]

\hline

\multicolumn{5}{|l|}{\textbf{\small Quantization}} \\
\hline
Incremental Quantization \cite{chen2021quantization}&
VGG-16, Deep Speech, 34-layer FCN &
Unspecified  &
Image Classification, ASR, Biomedical Image Segmentation &
Memory reductions between 3.5x and 6.4x \\[2mm]

QuantEdge \cite{mahmudov2025quantedge} &
YOLOv5m, DeepSpeech2, LSTM &
Jetson AGX Xavier \& Cluster, Asus Tinker Edge T, RB Pi 4, Dell T5820 WS  &
Object Detection, ASR, Driver Profiling &
lowest power consumption of comparative quantization methods \\[2mm]

Post-Training Quantization (PTQ) \cite{rachmanto2024characterizing}&
MobileNetV3-S/L, EfficientNet-B1/B3, DenseNet-169/201  &
Jetson Xavier NX, Jetson Orin Nano &
Image Classification &
67 and 55 percent lower inference latency; over 70 percent INT8 size reduction \\[2mm]

MobileQuant \cite{tan-etal-2024-mobilequant} &
TinyLlama-1.1B-Chat-v1.0, StableLM-2-1.6 and Gemma-2B  &
Snapdragon 8 Gen 3 HTP &
Large Language Models (LLMs) &
50 percent lower prompt-encoding power than a W8A16 baseline, measured on W8A8 TinyLlama-1.1B \\[2mm]

INT8 Deployment with Sub-INT8 Simulation \cite{kose2025bridging}&
TinyYOLOv2 &
STM32N6  &
Object Detection &
INT8 runs on-device in real time; INT4 and hybrid INT4/8 remain simulation-only for lack of native kernels \\[2mm]

Knowledge Distillation and INT8 Quantization \cite{hasan2025}&
4-model Ensemble distilled to ShuffleNetV2&
iOS/Android App (ONNX Runtime Mobile)  &
Leaf Disease Detection &
Ensemble teacher at 99.15\% accuracy; INT8 student retains 97.46\%\\[2mm]
\hline

\multicolumn{5}{|l|}{\textbf{\small Tensor Decomposition and Hardware-Aware Compression}} \\
\hline
Deep Learning Model Compression With Rank Reduction \cite{dai2023deep} &
VGG-16, ResNet-50, EfficientNet-B5 &
Raspberry Pi 4, Intel i7, NVIDIA RTX 3090 &
Image Classification &
10.41$\times$ storage reduction; 6.29$\times$ speedup; 13.96$\times$ FL overhead reduction \\[2mm]

Energy-Aware AI-Driven Framework \cite{zawish2022energy} &
AlexNet, VGG-16, ResNet-18, ResNet-32 &
Simulated solar-powered IoT nodes (Tesla K20 GPU) &
Image Classification &
75.3\% energy reduction with 6.3\% accuracy loss (AlexNet); 40\% with 0.6\% loss (VGG-16); 43\% with 0.45\% gain (ResNet-18) \\[2mm]

Doping \cite{thakker2021doping} &
LSTM: Medium LM, Large LM, RHN, GNMT &
Raspberry Pi 4 (Arm Cortex-A72) &
Language Modeling, Machine Translation &
10--25$\times$ compression, 2.5--5.5$\times$ inference speedup \\[2mm]
HALOC \cite{xiao2023haloc} &
ResNet-20, VGG-16, ResNet-18, MobileNetV2 &
NVIDIA Tesla V100, Jetson TX2, ASIC Eyeriss &
Image Classification &
0.9\% accuracy gain over baseline, 66.16\% FLOPs reduction (ResNet-18), 0.66\% gain over the prior best (MobileNetV2)  \\[2mm]

Hardware-Aware DNN Compression  \cite{zhang2025hardware} &
ResNet-50, MobileNetV1, VGG16, YOLOv8n &
NVIDIA Jetson Xavier NX, Jetson Nano &
Image Classification, Object Detection &
2.86$\times$ speedup on ResNet-50 at 1.0G FLOPs, lowest average latency across all device clusters \\[2mm]
Hardware-Aware Analog In-Memory Computing \cite{rasch2023hardware} &
ResNet-18/50, DenseNet-121, BERT, LSTM, RNN-T &
PCM-based Analog Crossbar Arrays (512$\times$512 tiles) &
Image Classification, NLP, Speech-to-Text &
5/11 DNNs reach iso-accuracy ($>$99\%); worst-case accuracy gap reduced from 21.81\% to 2.65\%  \\[2mm]

\hline

\multicolumn{5}{|l|}{\textbf{\small Knowledge Distillation}} \\
\hline
XAI-driven Knowledge Distillation \cite{cantini2024} &
Transformer / Bi-LSTM &
Not evaluated on edge hardware &
NLP Classification &
127$\times$ compression, 8.7$\times$ speedup. Uses local XAI explanations to guide distillation from a Transformer into a self-explainable Bi-LSTM. \\[2mm]

KD \& INT8 Quantization \cite{paparounas2024} &
ResNet-101 / MobileNet v3 &
Jetson Nano (CPU only) &
Image Classification &
44.8 ms latency at 3 Watts. Combines teacher-student distillation with post-training quantization for fast, low-power edge inference. \\[2mm]

Adaptive Knowledge Distillation \cite{wang2024} &
Multi-branch CNN &
Jetson Xavier NX &
Fault Diagnosis &
96.58\% memory reduction. Dynamically adjusts the distillation temperature to enhance cloud-to-edge knowledge transfer. \\[2mm]

Contrastive Representation Distillation \cite{setyanto2025} &
YOLOv4 / MobileNetV2 &
Jetson Orin Nano / RasPi 4B &
Object Detection &
6$\times$ smaller size; 2.6$\times$ faster on Orin Nano, 5.8$\times$ on RasPi 4B. Replaces the heavy CSPDarknet backbone with a distilled lightweight feature extractor for YOLOv4. \\[2mm]

Channel-wise KD \& Masked Generative Distillation \cite{saltik2025} &
YOLO11x / YOLO11n &
Jetson Orin Nano / RasPi 5 &
Weed Detection &
+2.5\% mAP50 without size increase. Transfers spatial attention maps from a heavy teacher to a compact student for precision agriculture. \\[2mm]

KD \& INT8 Quantization \cite{hasan2025} &
Ensemble CNN / ShuffleNetV2 &
Smartphones / Mobile App &
Leaf Disease Detection &
671$\times$ compression vs the ensemble teacher, 43.6$\times$ server-side speedup. Distills a four-model ensemble into an ultra-lightweight INT8 mobile network. \\[2mm]

\hline
\end{longtable}
\endgroup

\subsection{Quantization}
Quantization reduces the numerical precision used to represent weights, activations, or both: typically from a 32-bit floating point to 8 or 4-bit integers while leaving the network architecture and topology unchanged. Because parameter storage scales linearly with bit-width, lowering precision directly shrinks the memory footprint and bandwidth that dominate edge cost: a 7B-parameter model needs 28~GB in FP32 but only 7~GB in INT8 and 3.5~GB in INT4, which is often what makes a large model tractable on an Artificial Intelligence of Things (AIoT) device at all. Integer arithmetic also runs faster and at lower energy on the ARM and DSP units common in embedded hardware. The mapping is an affine transform   $q=\mathrm{clip}(\lfloor r/S\rceil+Z,\,q_{\min},\,q_{\max})$ \cite{jacob2018quantization} with scale $S$ and zero-point $Z$; post-training quantization (PTQ) applies it to a trained model, whereas quantization-aware training (QAT) simulates it during fine-tuning to recover accuracy.

Quantization alone already delivers measured gains. Incremental quantization reduces memory by 3.5--6.4$\times$ across image classification, speech recognition, and biomedical image segmentation models \cite{chen2021quantization}, and a characterization of post-training quantization across six classification CNNs reports inference latency reductions of 67\% on a Jetson Xavier NX and 55\% on an Orin Nano, with INT8 shrinking model size by more than 70\% \cite{rachmanto2024characterizing}. QuantEdge extends the comparison across heterogeneous targets, including Jetson AGX Xavier, Asus Tinker Edge T, and Raspberry Pi 4, and achieves the lowest power consumption among the quantization methods it evaluates \cite{mahmudov2025quantedge}. At the platform extremes, a microcontroller study runs INT8 TinyYOLOv2 in real time on an STM32N6 but finds INT4 and hybrid INT4/8 schemes blocked by missing native kernels, leaving sub-INT8 gains simulation-only \cite{kose2025bridging}, and MobileQuant moves quantized LLM inference onto phone hardware, halving prompt-encoding power relative to a W8A16 baseline for a W8A8 TinyLlama-1.1B on a Snapdragon 8 Gen 3 HTP \cite{tan-etal-2024-mobilequant}.

In deployment studies, quantization also acts as the final stage of a compression pipeline. Singh et al. apply INT8 quantization after pruning for a $10\times$ size reduction under TensorFlow Lite \cite{singh2023pruningquantization}, and Chang et al. fold 8-bit quantization into a mixed-pruning FPGA pipeline for $30\times$ compression \cite{chang2021mixedpruning}. Paired with distillation, static INT8 quantization lets a MobileNetV3 run at $44.8$~ms per image within a $3$~W budget on Jetson Nano \cite{paparounas2024}, while Hasan et al. distill a four-model ensemble into a ShuffleNetV2 student and quantize it to INT8, compressing $671\times$ relative to the ensemble at $97.46\%$ accuracy \cite{hasan2025}.  Quantization thus composes multiplicatively with pruning and distillation, but aggressive low-bit PTQ erodes accuracy when activation distributions are skewed, so the precision level must be matched to the target task and hardware.

\subsection{Tensor Decomposition and Hardware-Aware Compression}
Neural network tensors are stored as large multi-dimensional arrays and are often redundant containing more parameters than needed. Tensor decomposition exploits this by factorizing a large tensor into a few smaller components, reducing storage and computational cost without retraining from scratch. The most common form is Tucker decomposition, which approximates a large weight tensor as a small core tensor multiplied by per-mode factor matrices, $\mathcal{X}=\mathcal{G}\times_1 U_1\cdots\times_n U_n$ with core $\mathcal{G}$ and factor matrices $U_k$; truncating the multilinear rank below the original dimensions is what yields compression, and the convolution-oriented Tucker-2 variant factorizes only the channel modes. Related structured factorizations --- Kronecker product (KP) and low-rank matrix factorization for linear layers, optionally augmented by a sparse ``doping'' correction --- trade expressiveness for parameter count and help at the extreme ratios edge deployment demands.

How much of this theoretical saving reaches the device depends on how aggressively the rank is reduced, and on edge hardware these factorizations have mostly been validated on convolutional backbones for image classification and detection. The practical bottleneck is rank selection, because rank settings with very different FLOP counts often map to nearly identical on-device latency.

Deployment-oriented work closes this gap by making the rank choice explicit and hardware-guided. Dai et al.~\cite{dai2023deep} jointly learn an uncompressed model and its low-rank Tucker counterpart through iterative rank reduction, reaching up to $10.41\times$ storage reduction on VGG-16 across image classification tasks. The Hardware-Aware Automatic Low-rank Compression (HALOC) method casts rank selection as a differentiable, hardware-aware architecture search, cutting ResNet-18 FLOPs by $66.16\%$ while slightly improving accuracy on ImageNet \cite{xiao2023haloc}. Complementary structured methods optimize against measured device behavior: Homogeneous-Device Aware Pruning (HDAP) uses surrogate-guided pruning to minimize average latency across homogeneous device clusters for a $2.86\times$ speed-up on ResNet-50 on ImageNet~\cite{zhang2025hardware}, and an energy-aware reinforcement-learning scheme adapts compression depth to the harvested energy of IoT nodes for up to $75.3\%$ energy reduction on AlexNet--CIFAR-100 \cite{zawish2022energy}. Doped KP matrices compress LSTM models $10$--$25\times$ with a $2.5$--$5.5\times$ speed-up on a Raspberry Pi 4 across language modeling and machine translation tasks \cite{thakker2021doping}, while hardware-aware retraining restores iso-accuracy for analog in-memory accelerators under realistic noise and drift across image classification, natural language processing, and speech recognition workloads \cite{rasch2023hardware}. The recurring lesson is that FLOP reductions become real speed-ups only when the rank or structure is chosen against on-device measurements rather than proxy counts.

\subsection{Knowledge Distillation}
Knowledge distillation transfers the behavior of a large teacher into a compact student, training the student to match the teacher's soft outputs and, in richer variants, its intermediate features, attention maps, or explanations rather than hard labels alone, typically by minimizing loss $\mathcal{L}=\mathcal{L}_{CE}+\lambda\mathcal{L}_{KD}$, where $\mathcal{L}_{CE}$ is the standard hard-label loss and $\mathcal{L}_{KD}$ the teacher-matching term. Because it reduces parameter count while leaving numerical precision untouched, distillation is orthogonal to quantization, and the two compose multiplicatively. Feature- and relation-based variants such as contrastive, channel-wise, and masked generative distillation transfer spatial structure that plain logit matching misses, which matters for dense edge tasks like detection and segmentation.

On-device studies span language, vision, and industrial tasks. DiXtill guides distillation with integrated-gradient explanations, compressing a Transformer into a self-explainable Bi-LSTM at a $127\times$ ratio and $8.7\times$ speed-up for a $1.2\%$ accuracy drop \cite{cantini2024}. For detection, contrastive representation distillation replaces a heavy CSPDarknet backbone with a lightweight student in YOLOv4, giving a $6\times$ smaller model at $37.6$~ms per frame on a Jetson Orin Nano \cite{setyanto2025}, while channel-wise and masked generative distillation add $2.5\%$ mAP50 for weed detection at no extra cost \cite{saltik2025}. An ensemble of four CNN teachers distilled into a ShuffleNetV2 student holds $98.53\%$ accuracy on a unified leaf-disease benchmark, and $97.46\%$ after INT8 quantization \cite{hasan2025}, and an adaptive-temperature scheme in a cloud-edge pipeline cuts edge memory by $96.58\%$ for fault diagnosis on a Jetson Xavier NX \cite{wang2024}. Distillation followed by INT8 quantization recurs as a pattern: a distilled MobileNetV3 runs at $44.8$~ms within $3$~W on Jetson Nano \cite{paparounas2024}, and quantizing the distilled student pushes compression to $671\times$ relative to the ensemble teacher \cite{hasan2025}. The main cost is the centralized GPU training of heavy teachers, and compact students remain vulnerable to teacher bias and distribution drift in evolving edge environments.

\subsection{Deployment Guidelines}
The surveyed practice fixes five design choices for the empirical study:

\begin{itemize}
\item Prior edge compression studies use low bit formats like 8 bit or 4 bit. Following this practice, we evaluated \texttt{Q4\_K\_M}, our primary target for Raspberry Pi deployment, against other quantization levels such as \texttt{Q5\_K\_M} and \texttt{Q6\_K} in Table~\ref{tab:quant_prune_lora_results}.

\item Pruned SLMs and the image model are typically evaluated using compression ratio, model size and inference efficiency in FP16. This is why we report model level compression such as number of MLP channels removed and the deployment level results including inference speed in seconds per sample across GPU, CPU and Raspberry Pi in Table~\ref{tab:gpu_cpu_rpi_pruned_llm_results}.

\item After pruning, recovery is commonly performed using techniques like knowledge distillation, low-rank adaptation and fine tuning. We have used LoRA recovery after structured MLP pruning and evaluated how the models perform on QA datasets like SQuAD, Natural Questions and BoolQ.

\item Compression sensitivity differs across model architectures, so we compare three model families --- Qwen, Gemma, and TinyLlama --- under the same pruning, recovery, and quantization techniques.

\item For the semantic segmentation models, studies report how they are evaluated not only using model size but also task specific metrics after pruning, quantization and from our previous work of knowledge distillation. Following this, our study includes compressed image models like FPN~\cite{kirillov2019panoptic}, U-Net~\cite{ronneberger2015u}, and DeepLabV3+~\cite{chen2018deeplabv3plus} that have ResNet~\cite{he2016deep} and EfficientNet~\cite{tan2019efficientnet} backbones in Table~\ref{tab:1}.

\end{itemize}

\section{Empirical Study}
We empirically deploy several language and image segmentation models on GPU, CPU, and Raspberry Pi-based edge platforms. The detailed specifications of the testing platforms are provided in the following subsection, followed by result analysis and observations.

\begin{table*}[t]
\centering
\caption{Summary of the empirical experiments, reported results, and key deployment-level observations.}
\label{tab:experiment_summary}

\scriptsize
\renewcommand{\arraystretch}{1.15}
\setlength{\tabcolsep}{3pt}

\begin{tabularx}{\textwidth}{
    >{\raggedright\arraybackslash}p{0.14\textwidth}
    >{\raggedright\arraybackslash}X
    >{\raggedright\arraybackslash}p{0.20\textwidth}
    >{\raggedright\arraybackslash}p{0.27\textwidth}
}
\toprule

\textbf{Domain and Tasks}
&
\textbf{Experiments}
&
\textbf{Result References}
&
\textbf{Key Insights}
\\

\midrule

\textbf{LLMs}

\begin{itemize}[
    leftmargin=1.1em,
    itemsep=1pt,
    topsep=2pt,
    parsep=0pt,
    partopsep=0pt
]
    \item SQuAD
    \item BoolQ
    \item Natural Questions
\end{itemize}

&
Baseline benchmarking, DepGraph-MLP structured pruning, LoRA-based recovery,
and GGUF quantization of Gemma, Qwen, and TinyLlama models. The models are
evaluated on GPU, CPU, and Raspberry Pi platforms using task performance,
model size, inference latency, and tokens per second.

&
\begin{itemize}[
    leftmargin=1.1em,
    itemsep=1pt,
    topsep=2pt,
    parsep=0pt,
    partopsep=0pt
]
    \item Fig.~\ref{fig:model_size_param_speed}
    \item Fig.~\ref{fig:squad_boolq_performance}
    \item Fig.~\ref{fig:tps}
    \item Fig.~\ref{fig:phase_split}
    \item Table~\ref{tab:gpu_cpu_rpi_pruned_llm_results}
    \item Table~\ref{tab:quant_prune_lora_results}
    \item Table~\ref{tab:boolq_prior_collapse}
\end{itemize}

&
\begin{itemize}[
    leftmargin=1.1em,
    itemsep=1pt,
    topsep=2pt,
    parsep=0pt,
    partopsep=0pt
]
    \item Comprehensive deployment results for LLMs under different compression techniques.
    \item Pruning-induced forgetfulness and pruning-induced latency.
    \item Prefill- and decode-level analysis of LLM inference.
    \item BoolQ prior collapse under model compression.
\end{itemize}

\\

\midrule

\textbf{Image Models}

\begin{itemize}[
    leftmargin=1.1em,
    itemsep=1pt,
    topsep=2pt,
    parsep=0pt,
    partopsep=0pt
]
    \item Semantic segmentation
\end{itemize}

&
Baseline benchmarking, structured pruning, quantization, and
knowledge-distillation-assisted recovery of FPN, U-Net, and DeepLabV3+
models with ResNet-50 and EfficientNet-B3 backbones. The models are evaluated
using parameters, MACs, model size, mIoU, accuracy, wall latency, CPU latency,
and device temperature.

&
\begin{itemize}[
    leftmargin=1.1em,
    itemsep=1pt,
    topsep=2pt,
    parsep=0pt,
    partopsep=0pt
]
    \item Table~\ref{tab:1}
    \item Table~\ref{tab:post-dp}
\end{itemize}

&
\begin{itemize}[
    leftmargin=1.1em,
    itemsep=1pt,
    topsep=2pt,
    parsep=0pt,
    partopsep=0pt
]
    \item Comprehensive deployment results for image models under different compression techniques.
    \item Limitations of PyTorch's default quantization workflow for convolution-dominated segmentation models.
    \item Pruning-induced performance degradation and knowledge-distillation-assisted recovery.
    \item Task performance is not monotonic with model size or parameter count.
\end{itemize}

\\

\bottomrule
\end{tabularx}
\end{table*}

\begin{table*}[t]
\centering
\caption{GPU, CPU, and Raspberry Pi inference results for DepGraph-MLP (DG-MLP) pruning and LoRA-based recovery. Pruning shrinks the FP16 checkpoint yet slows inference on every internally controlled CPU comparison, and LoRA recovery does not restore baseline quality. All task metrics and latencies are measured with \texttt{llama.cpp} serving GGUF artifacts (\texttt{Q4\_K\_M} for the primary platform comparison); the FP16 column reports the pre-quantization artifact size, not the inference precision.}
\label{tab:gpu_cpu_rpi_pruned_llm_results}
\scriptsize
\renewcommand{\arraystretch}{0.95}
\setlength{\tabcolsep}{1.7pt}

\begin{adjustbox}{max width=\textwidth}
\begin{tabular}{@{}lllccrrrrrrrrrrrr@{}}
\toprule
\textbf{Model} &
\textbf{Method} &
\makecell{\textbf{Target/Actual}\\\textbf{Pruning (\%)}} &
\makecell{\textbf{FP16}\\\textbf{Size (MB)}} &
\makecell{\textbf{MLP Channels}\\\textbf{Removed}} &
\multicolumn{4}{c}{\textbf{GPU Results}} &
\multicolumn{4}{c}{\textbf{CPU Results}} &
\multicolumn{4}{c}{\textbf{Raspberry Pi Results}} \\
\cmidrule(lr){6-9}
\cmidrule(lr){10-13}
\cmidrule(lr){14-17}
&
&
&
&
&
\makecell{\textbf{SQuAD}\\\textbf{F1}} &
\makecell{\textbf{BoolQ}\\\textbf{Strict}} &
\makecell{\textbf{NQ F1}\\\textbf{/ EM}} &
\makecell{\textbf{Speed}\\\textbf{(s/sample)}} &
\makecell{\textbf{SQuAD}\\\textbf{F1}} &
\makecell{\textbf{BoolQ}\\\textbf{Strict}} &
\makecell{\textbf{NQ F1}\\\textbf{/ EM}} &
\makecell{\textbf{Speed}\\\textbf{(s/sample)}} &
\makecell{\textbf{SQuAD}\\\textbf{F1}} &
\makecell{\textbf{BoolQ}\\\textbf{Parsed}} &
\makecell{\textbf{NQ F1}\\\textbf{/ EM}} &
\makecell{\textbf{Speed}\\\textbf{(s/sample)}} \\
\midrule

\multicolumn{17}{@{}l}{\textbf{Gemma 3 1B IT}} \\
\midrule
Gemma-1B & Baseline & 0/0.00 & 1913.62 & 0   & 64.99 & 53 & 9.11/1 & 0.1167 & 64.01 & 32 & 8.39/1 & 0.5921 & 82.26 & 69 & 11.70/9 & 8.48 \\
Gemma-1B & DG-MLP & 1/1.04 & 1901.28 & 72  & 0.50 & 0  & 2.27/0 & 0.1078 & 0.83 & 0 & 1.48/0 & 0.8456 & 0.95 & 100.00 & 1.96/0 & 12.77 \\
Gemma-1B & DG-MLP & 3/3.01 & 1877.97 & 208 & 0.00 & 0  & 0.00/0 & 0.1165 & 0.50 & 0 & 0.50/0 & 0.6500 & 1.33 & 0.00 & 0.17/0 & 12.45 \\
Gemma-1B & DG-MLP & 5/5.03 & 1853.98 & 348 & 0.29 & 0  & 0.00/0 & 0.1163 & 0.00 & 0 & 0.00/0 & 0.7465 & 0.25 & 0.00 & 0.00/0 & 8.88 \\
Gemma-1B & DG-MLP & 7/7.00 & 1830.67 & 484 & 0.00 & 0  & 0.00/0 & 0.1167 & 0.20 & 0 & 0.00/0 & 0.7343 & 0.00 & 0.00 & 0.45/0 & 9.22 \\

Gemma-1B & DG-MLP+LoRA & 1/1.04 & 1901.28 & 72  & 13.59 & 1 & 3.52/0 & 0.0940 & 13.74 & 1 & 3.42/0 & 0.7571 & 1.57 & 0.00 & 1.83/0 & 20.59 \\
Gemma-1B & DG-MLP+LoRA & 3/3.01 & 1877.97 & 208 & 0.14 & 0 & 1.03/0 & 0.0962 & 1.89 & 0 & 0.64/0 & 0.6038 & 0.52 & 0.00 & 0.79/0 & 8.57 \\
Gemma-1B & DG-MLP+LoRA & 5/5.03 & 1853.98 & 348 & 0.22 & 1 & 0.65/0 & 0.0968 & 0.63 & 0 & 0.75/0 & 0.7355 & 0.38 & 0.00 & 0.45/0 & 8.81 \\
Gemma-1B & DG-MLP+LoRA & 7/7.00 & 1830.67 & 484 & 0.00 & 0 & 0.53/0 & 0.0967 & 0.37 & 0 & 0.39/0 & 0.7333 & 0.31 & 0.00 & 0.00/0 & 9.25 \\

\midrule
\multicolumn{17}{@{}l}{\textbf{Qwen3.5 0.8B}} \\
\midrule
Qwen3.5 & Baseline & 0/0.00 & 1446.48 & 0   & 84.08 & 73 & 7.69/2 & 0.1567 & 86.30 & 70 & 8.18/2 & 0.7826 & 87.93 & 66.67 & 10.22/4 & 7.53 \\
Qwen3.5 & DG-MLP   & 1/1.00 & 1441.42 & 36  & 64.56 & 57 & 8.78/4 & 0.1775 & 68.49 & 52 & 8.01/4 & 0.9150 & 49.08 & 64.13 & 4.39/0 & 12.75 \\
Qwen3.5 & DG-MLP   & 3/3.01 & 1431.29 & 108 & 44.63 & 36 & 5.43/3 & 0.2199 & 40.03 & 36 & 4.44/2 & 0.9955 & 24.59 & 55.26 & 3.31/1 & 25.84 \\
Qwen3.5 & DG-MLP   & 5/5.02 & 1421.17 & 180 & 23.59 & 31 & 4.30/2 & 0.2590 & 26.14 & 31 & 6.32/3 & 1.1076 & 11.91 & 48.89 & 2.08/0 & 19.93 \\
Qwen3.5 & DG-MLP   & 7/7.03 & 1411.04 & 252 & 21.63 & 44 & 4.63/1 & 0.3415 & 23.08 & 40 & 3.99/1 & 1.1856 & 10.07 & 46.88 & 2.24/0 & 12.08 \\

Qwen3.5 & DG-MLP+LoRA & 1/1.00 & 1441.42 & 36 & 51.27 & 72 & 5.16/1 & 0.1557 & 53.32 & 72 & 6.39/1 & 0.9013 & 51.19 & 74.49 & 6.16/1 & 8.07 \\
Qwen3.5 & DG-MLP+LoRA & 3/3.01  & 1431.29 & 108 & 59.34 & 34 & 7.28/4 & 0.1626 & 65.51 & 34 & 6.76/3 & 0.9410 & 61.04 & 37.11 & 8.57/5 & 8.56 \\
Qwen3.5 & DG-MLP+LoRA & 5/5.02  & 1421.17 & 180 & 54.46 & 60 & 3.24/1 & 0.1909 & 54.45 & 56 & 2.81/1 & 1.0264 & 48.98 & 53.68 & 4.37/1 & 21.51 \\
Qwen3.5 & DG-MLP+LoRA & 7/7.03  & 1411.04 & 252 & 42.97 & 71 & 6.26/3 & 0.2163 & 41.86 & 65 & 4.55/2 & 1.1518 & 42.58 & 69.07 & 5.66/2 & 10.78 \\
Qwen3.5 & DG-MLP+LoRA & 10/10.04 & 1395.86 & 360 & 33.70 & 62 & 4.24/0 & 0.2590 & 34.52 & 64 & 4.74/1 & 1.2924 & 36.23 & 65.31 & 4.34/1 & 12.27 \\
\midrule
\multicolumn{17}{@{}l}{\textbf{TinyLlama 1.1B Chat}} \\
\midrule
TinyLlama & Baseline & 0/0.00 & 2099.06 & 0   & 57.69 & 66 & 9.90/2 & 0.1100 & 55.15 & 66 & 6.50/1 & 0.4504 & 18.84 & 72.09 & 6.63/1 & 10.72 \\
TinyLlama & DG-MLP   & 1/1.07 & 2083.59 & 60  & 9.55  & 26 & 3.24/0 & 0.0759 & 10.41 & 27 & 3.15/0 & 0.8916 & 13.81 & 68.83 & 4.58/0 & 14.35 \\
TinyLlama & DG-MLP   & 3/3.05 & 2054.71 & 172 & 13.20 & 32 & 2.82/0 & 0.0728 & 16.35 & 35 & 3.00/0 & 0.8848 & 10.34 & 66.67 & 4.09/0 & 14.94 \\
TinyLlama & DG-MLP   & 5/5.04 & 2025.84 & 284 & 7.15  & 22 & 4.67/0 & 0.0767 & 7.56  & 24 & 2.77/0 & 0.8873 & 5.11  & 76.92 & 3.63/0 & 15.04 \\
TinyLlama & DG-MLP   & 7/7.03 & 1996.96 & 396 & 4.39  & 8  & 2.49/0 & 0.0710 & 5.06  & 10 & 3.27/0 & 0.8289 & 5.05  & 71.79 & 2.91/0 & 14.83 \\

TinyLlama & DG-MLP+LoRA & 1/1.07 & 2083.59 & 60  & 36.59 & 11 & 4.57/0 & 0.0717 & 39.44 & 9  & 3.57/0 & 0.8065 & 5.78  & 59.18 & 2.93/0 & 15.60 \\
TinyLlama & DG-MLP+LoRA & 3/3.05 & 2054.71 & 172 & 32.79 & 18 & 5.13/1 & 0.0695 & 33.30 & 18 & 5.19/1 & 0.8070 & 11.47 & 42.59 & 2.57/0 & 18.80 \\
TinyLlama & DG-MLP+LoRA & 5/5.04 & 2025.84 & 284 & 27.74 & 42 & 4.51/0 & 0.0789 & 28.65 & 43 & 3.72/0 & 0.8247 & 13.69 & 46.00 & 2.84/0 & 15.47 \\
TinyLlama & DG-MLP+LoRA & 7/7.03 & 1996.96 & 396 & 27.84 & 13 & 4.38/0 & 0.0756 & 29.74 & 13 & 3.19/0 & 0.8227 & 13.33 & 64.58 & 3.11/0 & 15.28 \\
\midrule

\bottomrule
\end{tabular}
\end{adjustbox}

\vspace{1mm}
\begin{minipage}{0.98\textwidth}
\footnotesize
\textit{Note:} Raspberry Pi results use parsed BoolQ accuracy because the raw strict score is sensitive to output-format parsing failures; a parsed accuracy of 100 for a collapsed pruned variant reflects one or a few parseable responses rather than genuine competence. Speed is reported in seconds per sample. The Raspberry Pi baseline and pruned rows come from separate evaluation campaigns whose prompt templates are not controlled against each other, so Raspberry Pi baseline-to-pruned comparisons carry a prompting confound; the GPU and CPU columns are each internally controlled, except that TinyLlama's GPU baseline comes from the quantization-sweep harness rather than the campaign that produced its pruned rows.
\end{minipage}
\end{table*}
\subsection{Experimental Setups}

For the language model experiments, we used an NVIDIA RTX 4070 laptop GPU, and to finetune image segmentation models, we used NVIDIA L40 GPUs with 48 GB memory each. The CPU-based experiments were conducted on an AMD Ryzen 9 7940HS processor running 64-bit Ubuntu Linux, with PyTorch 2.12 Stable and Python 3.11. For edge-device evaluation, we used a Raspberry Pi 4B equipped with a 64-bit quad-core Arm Cortex-A72 processor, 8 GB LPDDR4 RAM, and 128 GB storage. The Raspberry Pi runs Ubuntu 26.04 LTS with Torchvision and other required packages, as specified in the dependency list of our project repository. All LLM task metrics and latencies reported in this study were measured with \texttt{llama.cpp} serving GGUF artifacts (\texttt{Q4\_K\_M} for the primary platform comparison) on every platform; PyTorch was used for structured pruning, LoRA recovery, and the image-model experiments. The image-model deployment results carried over from our prior work were measured on a separate Raspberry Pi 4B with 2 GB LPDDR4 RAM running Ubuntu 22.04 LTS ARM64 with Python 3.10 and PyTorch 2.1.0~\cite{Our_CCNC_2026work}.

\subsection{Applied Compression Techniques}
Here we describe the specific compression techniques applied in our empirical study; the general formulation of each family is given in Section~\ref{sec:overview}, so here we focus on the configuration choices particular to our edge deployment.
\subsubsection{DepGraph Pruning}

Structured pruning changes the topology of the neural flow graph, while quantization preserves topology but changes numerical precision. Their effectiveness can be compared by measuring graph compression, output drift, task accuracy, and hardware efficiency together.

Removing one feed-forward channel is not a local edit: it forces the removal of every weight slice coupled to it. DepGraph \cite{fang2023depgraph} makes that coupling explicit, so we partition the parameters $\Theta$ of \eqref{eq:pruning_optimization} into $m$ dependency groups $\mathcal{G}=\{g_1,\dots,g_m\}$, where all weight slices in a group $g$ share one prunable dimension and must be pruned jointly. Each prunable dimension $k$ of a group is scored by the aggregated squared norm of its coupled slices,
\begin{equation}
I_{g,k}=\sum_{w\in g}\lVert w[k]\rVert_2^{2},
\label{eq:depgraph_group_importance}
\end{equation}
where the sum runs over the weight slices $w$ coupled in group $g$, $w[k]$ is the portion of each slice indexed by the shared dimension $k$, and $\lVert\cdot\rVert_2$ is the Euclidean norm; the score $I_{g,k}$ therefore reflects the joint magnitude of every parameter the dimension touches rather than any layer in isolation.

We restrict pruning to the MLP width of each transformer block, whose dependency group couples the gate, up, and down projections through their shared intermediate channel. A pruning ratio $p\in[0,1]$ removes the channels with the smallest $I_{g,k}$, reducing the original feed-forward width $d_{\mathrm{ffn}}$ to the pruned width
\begin{equation}
d' = d_{\mathrm{ffn}} - \lfloor p\, d_{\mathrm{ffn}} \rfloor ,
\label{eq:pruned_width}
\end{equation}
where $\lfloor\cdot\rfloor$ rounds down to an integer channel count, realizing the mask $M$ of \eqref{eq:pruning_optimization} at channel granularity while leaving attention and embedding parameters intact.

A ratio that is optimal for the dense model, however, need not remain optimal once the pruned model is quantized for deployment. Block-wise $k$-quant GGUF formats store weights in super-blocks of $Q_K=256$ elements, so a pruned width is stored at the target precision only when it stays aligned to the block size,
\begin{equation}
d' \equiv 0 \pmod{Q_K},
\label{eq:kquant_alignment}
\end{equation}
and otherwise the affected tensors fall back to a higher-precision format, so that the deployed model can grow rather than shrink.

\subsubsection{Low-Rank Recovery (LoRA)}

To repair the accuracy lost to pruning we use a low-rank update rather than full fine-tuning, the deployment-friendly member of the tensor-decomposition family reviewed earlier. For a pruned weight matrix $W\in\mathbb{R}^{d\times k}$ that is kept frozen, LoRA \cite{hu2021lora} learns a correction constrained to a rank-$r$ factorization
\begin{equation}
W' = W + \Delta W,\qquad \Delta W = BA,
\label{eq:lora_update}
\end{equation}
with $B\in\mathbb{R}^{d\times r}$, $A\in\mathbb{R}^{r\times k}$, and rank $r\ll\min(d,k)$, so that only $r(d+k)$ parameters are trained instead of the $dk$ of a dense update. The recovery reuses the same low-rank principle as Tucker and low-rank matrix factorization, but applies it to the correction signal rather than to the weights themselves, keeping the deployed model dense while the adaptation cost stays small.

Every recovery run instantiates this update identically across all four evaluated models: rank-$8$ adapters ($\alpha=16$, dropout $0.05$) attach to exactly the three pruned MLP projections (gate, up, and down) with the base weights frozen, leaving $0.35$--$0.50\%$ of parameters trainable. Each adapter trains for $100$ optimizer steps at a learning rate of $2\times10^{-4}$ with an effective batch of eight $256$-token sequences drawn from WikiText-2, so recovery uses a generic language-modeling corpus and the downstream QA benchmarks are never seen during fine-tuning.

\subsubsection{Quantization}

Unlike pruning, quantization preserves the topology and instead lowers the precision of the stored weights through the affine mapping introduced in Section~\ref{sec:overview}; for deployment what matters is how locally its scale and zero-point are allowed to vary.

We deploy with the block-wise $k$-quant GGUF formats of \texttt{llama.cpp}, which assign $S$ and $Z$ per super-block of $Q_K=256$ weights from~\eqref{eq:kquant_alignment} rather than per tensor. Each super-block is split into sub-blocks whose scales are themselves quantized against a single super-block scale, giving the two-level encoding
\begin{equation}
\hat{w}_{b,i}=S_b\,q_{b,i}-m_b,\qquad S_b=\Delta\,\tilde{s}_b,
\label{eq:kquant_blockscale}
\end{equation}
where $q_{b,i}$ is the low-bit code of weight $i$ in sub-block $b$, $\tilde{s}_b$ and $m_b$ its quantized scale and offset, and $\Delta$ the shared super-block step. This hierarchy concentrates the bit budget on the weight codes while still adapting the scale locally, and each format is summarized by its effective bits per weight (bpw), which fixes the on-disk size $\text{Size}\approx N_{\text{params}}\,\text{bpw}/8$. The variants we evaluate---\texttt{Q8\_0}, \texttt{Q6\_K}, \texttt{Q5\_K\_M}, and \texttt{Q4\_K\_M}---form a precision ladder from roughly $8.5$ down to under $5$ bpw, with \texttt{Q4\_K\_M} taken as the primary memory-constrained target.

\subsubsection{Knowledge Distillation}

Knowledge distillation, whose formulation is given in Section~\ref{sec:overview}, was applied to the image models in our prior work \cite{Our_CCNC_2026work}; for the LLM pipeline here, LoRA fills the post-pruning recovery role instead.

\subsection{Experiments}

Table~\ref{tab:experiment_summary} summarizes the experiments conducted in this empirical study. These experiments measure how task quality, artifact size, and latency move together across compression methods and hardware, exposing deployment-level behavior that aggregate performance reports hide.

Instruction formatting moves extractive QA scores independently of the model, so we additionally ran a prompt sweep to calibrate that effect. For Gemma 3 1B IT, Qwen2.5 1.5B Instruct (an auxiliary fourth model), and TinyLlama 1.1B Chat in \texttt{Q4\_K\_M}, we evaluated four instruction templates over the same 100 SQuAD items on both GPU and CPU under greedy decoding, varying only the template. The sweep gives the score range a template change alone can produce on fixed hardware, which we use below to separate prompting effects from platform effects.

\begin{figure*}[!t]
    \centering
    \includegraphics[
        width=\textwidth,
        height=0.4\textheight,
        keepaspectratio
    ]{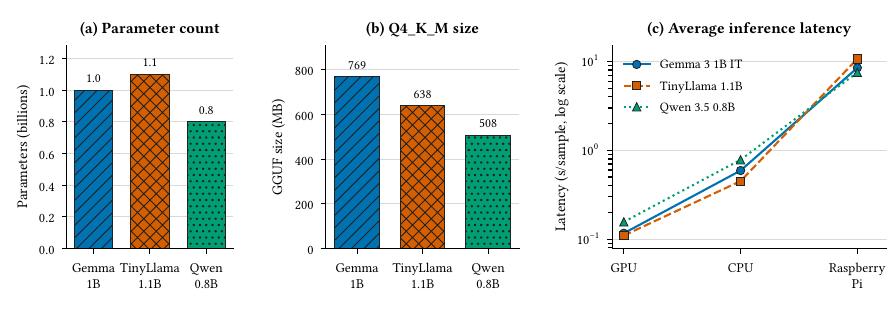}
    \caption{Comparison of parameter count, Q4\_K\_M GGUF model size, and average fixed-prompt inference latency (the mean of per-sample latencies over the SQuAD, BoolQ, and NQ runs) for Gemma 3 1B IT, TinyLlama 1.1B, and Qwen3.5 0.8B on GPU, CPU, and Raspberry Pi platforms.}
    \label{fig:model_size_param_speed}
\end{figure*}

\begin{figure*}[!t]
    \centering
    \includegraphics[
        width=\textwidth,
        height=0.4\textheight,
        keepaspectratio
    ]{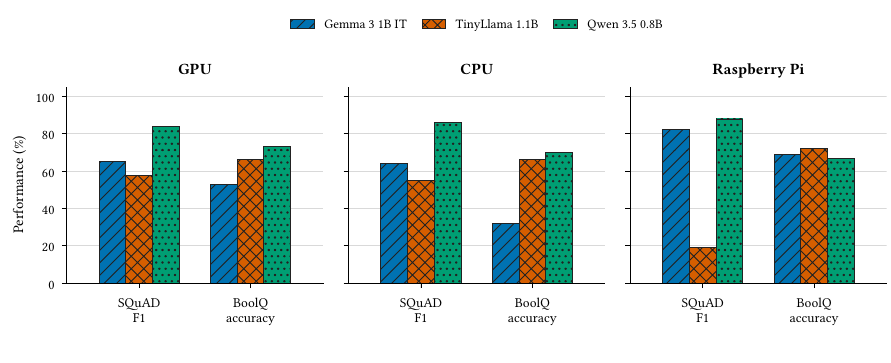}
    \caption{Fixed-prompt Q4\_K\_M baseline performance of small language models across GPU, CPU, and Raspberry Pi platforms using SQuAD F1 and BoolQ accuracy. GPU and CPU BoolQ values are strict accuracies (GPU parse rates: Gemma 78\%, TinyLlama 92\%, Qwen 100\%; CPU: 41\%, 93\%, 100\%); Raspberry Pi values are parsed accuracies (parse rates: 100\%, 86\%, and 99\%, respectively).}
    \label{fig:squad_boolq_performance}
\end{figure*}

\begin{figure*}[!htbp]
    \centering
    \includegraphics[width=0.8\textwidth]{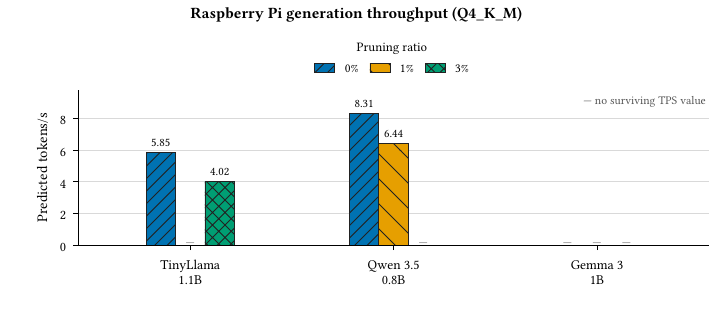}
    \caption{Raspberry Pi predicted-token throughput for Q4\_K\_M baselines (0\%) and surviving DepGraph-MLP structured-pruning runs. Dashes denote model--ratio cells without a surviving TPS value; Gemma has no baseline TPS because its coherent run did not capture a separate server TPS log.}
    \label{fig:tps}
\end{figure*}

\newcommand{\onehead}[1]{\makecell{\textbf{#1}\\[-0.3mm]\phantom{\textbf{X}}}}
\newcommand{\twohead}[2]{\makecell{\textbf{#1}\\[-0.3mm]\textbf{#2}}}

\begin{table*}[t]
\centering
\caption{Quantized Pruning and LoRA Recovery Results Across Models. BoolQ accuracy is computed over parseable responses; the collapsed pruned Gemma variants produce a parseable answer on only 1--4 of 100 prompts, so a BoolQ score of 100 reflects a single correct parseable response (strict accuracy 1\%); the pruned TinyLlama variants parse on only 12--48 of 100 prompts, so their BoolQ scores overstate competence for the same reason and two of them exceed the TinyLlama baseline, which parses on 92; Table~\ref{tab:boolq_prior_collapse} illustrates the same prior-collapse pathology in the Qwen family. Quality Mean is the unweighted mean of SQuAD F1, BoolQ accuracy, and NQ F1; for the collapsed Gemma variants it is dominated by the degenerate BoolQ term and should not be read as a quality ranking. The pruned and LoRA-recovered rows were independently re-evaluated in the quantization-sweep harness rather than reused from Table~\ref{tab:gpu_cpu_rpi_pruned_llm_results}; the screening and recovery campaigns used SQuAD/BoolQ generation ceilings of 12/3 tokens whereas this sweep used 48/16, so overlapping configurations are not expected to reproduce the earlier table exactly.}
\label{tab:quant_prune_lora_results}
\scriptsize
\setlength{\tabcolsep}{2.0pt}
\renewcommand{\arraystretch}{1.05}

\begin{tabular*}{\textwidth}{@{\extracolsep{\fill}}
C{1.25cm}
C{0.65cm}
C{0.72cm}
C{0.78cm}
C{0.88cm}
C{0.72cm}
C{0.62cm}
C{0.72cm}
C{0.62cm}
C{0.62cm}
C{0.78cm}
C{1.02cm}
@{}}
\toprule
\onehead{Characteristic} &
\twohead{Target}{Prune} &
\twohead{LoRA}{Recovery} &
\twohead{Quant.}{Level} &
\twohead{GGUF Size}{(MB)} &
\twohead{SQuAD}{F1} &
\twohead{SQuAD}{EM} &
\twohead{BoolQ}{Parsed} &
\twohead{NQ}{F1} &
\twohead{NQ}{EM} &
\twohead{GPU Speed}{(s/sample)} &
\twohead{Quality}{Mean} \\
\midrule

\multicolumn{12}{@{}l}{\textbf{Qwen3.5 0.8B}} \\
\midrule
\textbf{Baseline} & \textbf{0} & \textbf{N/A} & \textbf{Q8\_0} & \textbf{774.23} & \textbf{92.8} & \textbf{91} & \textbf{73} & \textbf{10.38} & \textbf{4} & \textbf{0.13} & \textbf{58.727} \\
\textbf{Baseline} & \textbf{0} & \textbf{N/A} & \textbf{Q6\_K} & \textbf{600.57} & \textbf{93.35} & \textbf{91} & \textbf{72} & \textbf{11.08} & \textbf{5} & \textbf{0.1233} & \textbf{58.81} \\
\textbf{Baseline} & \textbf{0} & \textbf{N/A} & \textbf{Q5\_K\_M} & \textbf{551.22} & \textbf{93.85} & \textbf{92} & \textbf{75} & \textbf{10.18} & \textbf{5} & \textbf{0.1233} & \textbf{59.677} \\
\textbf{Baseline} & \textbf{0} & \textbf{N/A} & \textbf{Q4\_K\_M} & \textbf{507.85} & \textbf{84.08} & \textbf{82} & \textbf{73} & \textbf{7.69} & \textbf{2} & \textbf{0.1567} & \textbf{54.923} \\

\addlinespace[0.45mm]

Pruned & 1 & No & Q6\_K & 696.59 & 76.51 & 65 & 69 & 8.04 & 4 & 0.1667 & 51.183 \\
Pruned & 1 & No & Q5\_K\_M & 653.05 & 77.92 & 67 & 62 & 10.23 & 6 & 0.17 & 50.05 \\
Pruned & 1 & No & Q4\_K\_M & 612.06 & 64.56 & 49 & 57.58 & 8.78 & 4 & 0.1567 & 43.64 \\

\addlinespace[0.45mm]

Pruned & 1 & Yes & Q6\_K & 696.59 & 57.66 & 45 & 73 & 6.08 & 3 & 0.1633 & 45.58 \\
Pruned & 1 & Yes & Q5\_K\_M & 653.05 & 62.15 & 55 & 75 & 5.96 & 3 & 0.1333 & 47.703 \\
Pruned & 1 & Yes & Q4\_K\_M & 612.06 & 51.27 & 35 & 72.73 & 5.16 & 1 & 0.16 & 43.053 \\

\midrule

\multicolumn{12}{@{}l}{\textbf{TinyLlama 1.1B Chat}} \\
\midrule
\textbf{Baseline} & \textbf{0} & \textbf{N/A} & \textbf{Q8\_0} & \textbf{1115.62} & \textbf{44.95} & \textbf{21} & \textbf{71.28} & \textbf{8.94} & \textbf{0} & \textbf{0.1667} & \textbf{41.723} \\
\textbf{Baseline} & \textbf{0} & \textbf{N/A} & \textbf{Q6\_K} & \textbf{861.56} & \textbf{44.92} & \textbf{21} & \textbf{71.28} & \textbf{8.22} & \textbf{0} & \textbf{0.1433} & \textbf{41.473} \\
\textbf{Baseline} & \textbf{0} & \textbf{N/A} & \textbf{Q5\_K\_M} & \textbf{745.82} & \textbf{42.78} & \textbf{17} & \textbf{71.88} & \textbf{7.99} & \textbf{0} & \textbf{0.1333} & \textbf{40.883} \\
\textbf{Baseline} & \textbf{0} & \textbf{N/A} & \textbf{Q4\_K\_M} & \textbf{637.81} & \textbf{57.69} & \textbf{40} & \textbf{71.74} & \textbf{9.9} & \textbf{2} & \textbf{0.11} & \textbf{46.443} \\

\addlinespace[0.45mm]

Pruned & 3 & No & Q6\_K & 1120.14 & 13.59 & 0 & 68.75 & 3.56 & 0 & 0.1733 & 28.633 \\
Pruned & 3 & No & Q5\_K\_M & 1023.89 & 13.65 & 0 & 70.21 & 4.47 & 0 & 0.16 & 29.443 \\
Pruned & 3 & No & Q4\_K\_M & 933.3 & 14.82 & 0 & 78.57 & 3.5 & 0 & 0.16 & 32.297 \\

\addlinespace[0.45mm]

Pruned & 1 & Yes & Q6\_K & 1137.66 & 33.31 & 18 & 67.86 & 5.69 & 1 & 0.1767 & 35.62 \\
Pruned & 1 & Yes & Q5\_K\_M & 1040.13 & 34.34 & 18 & 63.33 & 5.51 & 1 & 0.1733 & 34.393 \\
Pruned & 1 & Yes & Q4\_K\_M & 948.34 & 34.78 & 22 & 75 & 4.41 & 0 & 0.1467 & 38.063 \\

\midrule

\multicolumn{12}{@{}l}{\textbf{Gemma 3 1B IT}} \\
\midrule
\textbf{Baseline} & \textbf{0} & \textbf{N/A} & \textbf{Q8\_0} & \textbf{1019.77} & \textbf{59.31} & \textbf{44} & \textbf{71.43} & \textbf{7.14} & \textbf{0} & \textbf{0.13} & \textbf{45.96} \\
\textbf{Baseline} & \textbf{0} & \textbf{N/A} & \textbf{Q6\_K} & \textbf{964.87} & \textbf{58.74} & \textbf{44} & \textbf{72.04} & \textbf{7.64} & \textbf{2} & \textbf{0.1233} & \textbf{46.14} \\
\textbf{Baseline} & \textbf{0} & \textbf{N/A} & \textbf{Q5\_K\_M} & \textbf{811.91} & \textbf{56.66} & \textbf{40} & \textbf{74.47} & \textbf{8.26} & \textbf{1} & \textbf{0.1267} & \textbf{46.463} \\
\textbf{Baseline} & \textbf{0} & \textbf{N/A} & \textbf{Q4\_K\_M} & \textbf{768.72} & \textbf{64.99} & \textbf{53} & \textbf{67.95} & \textbf{9.11} & \textbf{1} & \textbf{0.1167} & \textbf{47.35} \\

\addlinespace[0.45mm]

Pruned & 1 & No & Q6\_K & 1189.3 & 1.43 & 0 & 100 & 3.56 & 2 & 0.2833 & 34.997 \\
Pruned & 1 & No & Q5\_K\_M & 1050.73 & 1.54 & 0 & 100 & 1.88 & 0 & 0.2667 & 34.473 \\
Pruned & 1 & No & Q4\_K\_M & 1020.14 & 0.2 & 0 & 100 & 2.12 & 0 & 0.24 & 34.107 \\

\addlinespace[0.45mm]

Pruned & 1 & Yes & Q6\_K & 1189.3 & 4.24 & 0 & 0 & 1.67 & 0 & 0.2067 & 1.97 \\
Pruned & 1 & Yes & Q5\_K\_M & 1050.73 & 4.14 & 1 & 25 & 2.58 & 0 & 0.1867 & 10.573 \\
Pruned & 1 & Yes & Q4\_K\_M & 1020.14 & 12.58 & 9 & 66.67 & 2.67 & 0 & 0.1567 & 27.307 \\

\bottomrule
\end{tabular*}
\end{table*}

\section{Domain-Specific Result Analysis}

\subsection{LLM}
In this study, we evaluate the deployment behavior of small-scale LLMs with approximately 1B parameters across three representative hardware platforms: GPU, CPU, and Raspberry Pi-based edge devices. Specifically, we benchmark Gemma 3 1B IT~\cite{gemmateam2025gemma3}, TinyLlama 1.1B~\cite{zhang2024tinyllama}, and Qwen3.5 0.8B~\cite{qwen2026qwen35} using standard extractive and binary question-answering metrics, including SQuAD F1, SQuAD Exact Match (EM), Natural Questions (NQ) F1, NQ EM, and BoolQ accuracy.
\subsubsection{Baseline Results}
Table~\ref{tab:gpu_cpu_rpi_pruned_llm_results} shows that Qwen3.5 0.8B provides the strongest overall performance, with the highest SQuAD F1 of 84.08 on GPU and 86.30 on CPU in the unpruned baseline. That advantage does not survive compression:
\begin{itemize}

    \item Pruning consistently reduces accuracy: for Qwen3.5, GPU SQuAD F1 falls as low as 21.63 from baseline of 84.08 at 7.03\% pruning, while CPU SQuAD F1 decreases from 86.30 to 23.08. LoRA recovery partially improves the pruned models, with the highest recovered SQuAD F1 for Qwen3.5 at 3.01\% pruning (40.03 to 65.51 on CPU) and the largest recovery gain for TinyLlama at 1.07\% pruning (10.41 to 39.44 on CPU). 
    \item In terms of latency, the inference has a wide range across different models and hardware. It spans from as fast as 0.0695 s/sample on GPU for the LoRA-recovered TinyLlama at 3\% pruning to 25.84 s/sample on Raspberry Pi for Qwen3.5 at 3\% pruning as shown in Table~\ref{tab:gpu_cpu_rpi_pruned_llm_results}. Model-size reduction is visible but moderate: Qwen3.5 decreases from 1446.48 MB to 1395.86 MB at 10.04\% pruning, TinyLlama decreases from 2099.06 MB to 1996.96 MB at 7.03\% pruning, and Gemma decreases from 1913.62 MB to 1830.67 MB at 7\% pruning. 
    \item Overall, the table shows that DepGraph-MLP pruning shrinks the pre-quantization FP16 checkpoint yet slows inference on every internally controlled CPU comparison, while the accuracy loss is substantial and LoRA recovery does not restore baseline performance.

\end{itemize}

For TinyLlama the uncontrolled prompting on Raspberry Pi runs against the measured degradation rather than toward it. Its Raspberry Pi baseline used \texttt{one\_word\_or\_phrase}, the weakest of the four templates in the sweep at 17.13 SQuAD F1, while its Raspberry Pi pruned runs used \texttt{no\_extra\_words}, the strongest at 55.29 --- a handicap of 38 points in the pruned models' favor. Every pruned Raspberry Pi score nonetheless falls below the 18.84 baseline, from 13.81 at $1\%$ pruning to 5.05 at $7\%$. The quality loss survives a confound pointed in its favor.

As shown in Fig.~\ref{fig:model_size_param_speed}, the evaluated small LLMs have comparable parameter counts and model sizes, ranging from 0.8B to 1.1B parameters and approximately 508 MB to 769 MB in Q4\_K\_M storage size. However, the average inference latency varies sharply across hardware platforms. For the three models evaluated on all platforms, i.e., Gemma 3 1B IT, TinyLlama 1.1B, and Qwen3.5 0.8B, the mean latency increases from 0.13 s/sample on GPU to 0.61 s/sample on CPU and 8.91 s/sample on Raspberry Pi. This corresponds to an average slowdown of approximately $4.8\times$ from GPU to CPU and roughly $70\times$ from GPU to Raspberry Pi. Model size therefore does not predict deployed latency: a $70\times$ spread opens across platforms at a similar 508--769~MB artifact size.

Figure~\ref{fig:squad_boolq_performance} shows that GPU and CPU execution yield broadly comparable SQuAD accuracy for the full-size models: the GPU-to-CPU gap in SQuAD F1 stays within three percentage points for all three models (64.99\% to 64.01\% for Gemma 3 1B IT, 57.69\% to 55.15\% for TinyLlama, and 84.08\% to 86.30\% for Qwen3.5). BoolQ separates the platforms more sharply: Gemma's strict accuracy falls from 53\% on GPU, where 78\% of its responses parse into a valid yes/no answer, to 32\% on CPU, where only 41\% do, while TinyLlama and Qwen3.5 remain within 66\%--73\% on both platforms. On Raspberry Pi, Qwen3.5 sustains a strong SQuAD F1 of 87.93\% under the same prompt template, generation settings, and evaluation items as its GPU and CPU rows, so its cross-platform trend is directly comparable. Gemma's 82.26\% and TinyLlama's 18.84\% come from campaigns whose prompt templates and prompting modes differ from those of their own GPU and CPU rows, so neither figure is a platform comparison. The prompt sweep shows how much that costs. On one GPU over the same 100 items, TinyLlama scores 55.29 SQuAD F1 under \texttt{no\_extra\_words}, the template behind its GPU and CPU rows, and 17.13 under \texttt{one\_word\_or\_phrase}, the template behind its Raspberry Pi baseline; a template change alone therefore reproduces 38.16 of the 38.85-point GPU-to-Raspberry-Pi difference without any change of hardware. Gemma's templates span 62.42 to 48.70 over the same items, a range that runs opposite to its apparent Raspberry Pi gain.

\subsubsection{Pruning Results}
Structured pruning removes parameters, yet every pruned variant in Table~\ref{tab:gpu_cpu_rpi_pruned_llm_results} runs slower than its unpruned baseline on the internally controlled CPU comparisons. Qwen3.5 rises from 0.7826 to 1.1856 s/sample on CPU at 7\% pruning and from 7.53 to as much as 25.84 s/sample on the Raspberry Pi, while TinyLlama nearly doubles its CPU latency at a pruning ratio of only 1\%. The Raspberry Pi baseline and pruned measurements come from separate campaigns whose prompt templates are not controlled against each other, so the Raspberry Pi ratio carries a prompting confound; the CPU comparison, which is controlled, moves in the same direction. The apparent contradiction resolves once per-sample latency is separated into its two factors: the number of tokens generated per sample, and the cost of each token. Pruning drives both upward.

The first factor grows because pruning damages output discipline before it removes knowledge: pruned models become more verbose and less format-compliant. The mean SQuAD response of Qwen3.5 lengthens from 29 characters at baseline to 61 characters at 7\% pruning, and its BoolQ responses more than double; these lengths are measured in the CPU fixed-prompt campaign under greedy decoding. On GPU, where memory bandwidth is plentiful, this accounts for essentially the entire slowdown -- Qwen's 2.2$\times$ latency increase closely tracks its 2.1$\times$ growth in response length, and Gemma, whose pruned outputs collapse to near-empty strings, shows no GPU slowdown at all. TinyLlama is excluded from this GPU comparison because its baseline row comes from the quantization-sweep harness rather than the campaign that produced its pruned rows, so its GPU latencies are not like-for-like.

The second factor is the size of the deployable artifact. Because DepGraph removes individual MLP channels, the pruned FFN dimensions no longer align with the $256$-element super-blocks of the k-quant GGUF format, and $24$ of $320$ tensors for Qwen3.5 fall back to higher-precision storage. The converted Q4\_K\_M weights therefore grow rather than shrink: from 507.85 to 612.06~MB for Qwen3.5 ($+21\%$), from 768.72 to 1020.14~MB for Gemma ($+33\%$), and from 637.81 to 948.34~MB for TinyLlama ($+49\%$). Since decoding on CPU- and Raspberry Pi-class devices is dominated by weight traffic, the larger artifact makes every generated token costlier, though not proportionally: token-weighted per-token decode time rises in all 18 matched baseline-to-pruned pairs of the quantization-sweep serving logs, by $0.5$--$30\%$ at \texttt{Q4\_K\_M} against byte growth of $21$--$49\%$, in the same cross-family order as the byte growth (Section~\ref{subsec:prefill_decode}). The LoRA-recovered TinyLlama variants, whose responses are only modestly longer than baseline yet run almost twice as slow on CPU, separate this byte-traffic effect from verbosity. We analyze the underlying prefill/decode and block-alignment mechanism in detail in Section~\ref{subsec:prefill_decode}.

\subsubsection{LoRA Recovery}
LoRA recovery in our runs is selective rather than universally restorative. LoRA was introduced as a parameter-efficient adaptation method \cite{hu2021lora}, QLoRA extends it to 4-bit finetuning \cite{dettmers2023qlora}, and LLM-Pruner applies LoRA-style post-training to restore pruned models \cite{ma2023llmpruner}; that line of work asks whether aggregate accuracy comes back. The per-task results in Table~\ref{tab:gpu_cpu_rpi_pruned_llm_results} show that recovery is uneven across task families. For TinyLlama at 1\% pruning, LoRA raises SQuAD F1 from 9.55 to 36.59 while BoolQ accuracy falls from 26 to 11. For Qwen3.5 at 1\% pruning the pattern inverts: BoolQ recovers from 57 to 72 while SQuAD F1 drops from 64.56 to 51.27 and NQ F1 from 8.78 to 5.16. At 5\% pruning LoRA lifts both SQuAD (23.59 to 54.46) and BoolQ (31 to 60) for Qwen3.5, yet NQ still degrades. Gemma recovers only partially (SQuAD F1 0.50 to 13.59) and remains effectively collapsed. An adapter can therefore make one metric look healthier while another silently gets worse: LoRA after structured pruning acts as a task-specific behavioral intervention, not a blanket restoration step, and recovery should be reported per task rather than as a single aggregate.

\begin{table*}[t]
\centering
\footnotesize
\setlength{\tabcolsep}{3pt}
\renewcommand{\arraystretch}{1.2}

\caption{Model comparison across Full, Quantized, and Structured Pruning (with Pruning Ratio=0.5) \cite{Our_CCNC_2026work}. Structured pruning reduces average parameters by 61.55\%, MACs by 43.14\%, and storage by 61.98\%, whereas default quantization leaves parameters and MACs unchanged and reduces storage by only 0.63\% on average.}
\label{tab:1}

\resizebox{\textwidth}{!}{%
\begin{tabular}{@{}%
l l
@{\hspace{8pt}} L{28mm}@{\hspace{4pt}}l@{\hspace{4pt}}l
@{\hspace{10pt}} L{28mm}@{\hspace{4pt}}l@{\hspace{4pt}}l
@{\hspace{10pt}} L{28mm}@{\hspace{4pt}}l@{\hspace{4pt}}l
@{}}
\hline
\multicolumn{1}{c}{\multirow{2}{*}{\textbf{Model}}} &
\multicolumn{1}{c}{\multirow{2}{*}{\textbf{Backbone}}} &
\multicolumn{3}{c}{\textbf{Full Size}} &
\multicolumn{3}{c}{\textbf{Quantized}} &
\multicolumn{3}{c}{\textbf{Structured Pruned (S=0.5)}} \\ \cline{3-11}
\multicolumn{1}{c}{} & \multicolumn{1}{c}{} &
\multicolumn{1}{c}{\scriptsize \textbf{Param / MAC / Size}} & \multicolumn{1}{c}{\textbf{mIoU}} & \multicolumn{1}{c}{\scriptsize \textbf{Mean Acc.(\%)}} &
\multicolumn{1}{c}{\scriptsize \textbf{Param / MAC / Size}} & \multicolumn{1}{c}{\textbf{mIoU}} & \multicolumn{1}{c}{\scriptsize \textbf{Mean Acc.(\%)}} &
\multicolumn{1}{c}{\scriptsize \textbf{Param / MAC / Size}} & \multicolumn{1}{c}{\textbf{mIoU}} & \multicolumn{1}{c}{\scriptsize \textbf{Mean Acc.(\%)}} \\ \hline\hline

FPN & ResNet-50
& 26.116 / 39.44 / 104.47 & 0.6162 & 69.48
& 26.116 / 39.44 / 104.27 & 0.6168 & 71.48
& 6.709  / 18.37 / 25.59  & 0.5784 & 64.61 \\

FPN & EfficientNet-B3
& 12.476 / 16.33 / 47.19 & 0.6728 & 75.59
& 12.476 / 16.33 / 47.00 & 0.6710 & 75.77
& 6.411  / 10.40 / 24.46  & 0.5252 & 61.22 \\

U-Net & ResNet-50
& 32.521 / 54.02 / 130.99 & 0.5612 & 61.91
& 32.521 / 54.02 / 130.88 & 0.5583 & 64.31
& 6.995 / 29.80 / 26.69   & 0.5517 & 62.86 \\

U-Net & EfficientNet-B3
& 13.159 / 19.51 / 49.93 & 0.6636 & 74.80
& 13.159 / 19.51 / 49.82 & 0.6603 & 74.79
& 6.896 / 14.93 / 26.31   & 0.4467 & 54.76 \\

DeepLabV3+ & ResNet-50
& 26.678 / 46.29 / 106.71 & 0.6560 & 73.85
& 26.678 / 46.29 / 106.51 & 0.6490 & 73.15
& 7.943 / 20.78 / 30.30   & 0.6258 & 70.97 \\

DeepLabV3+ & EfficientNet-B3
& 11.681 / 15.02 / 44.22  & 0.6611 & 74.13
& 11.681 / 15.02 / 43.02  & 0.6611 & 74.03
& 5.836  / 8.16  / 22.26   & 0.5360 & 63.72 \\

\hline
\end{tabular}
}
\end{table*}

\begin{table*}[ht]
\centering
\caption{Deployment statistics of segmentation models across different sparsity levels \cite{Our_CCNC_2026work}. Increasing the pruning ratio from 0.5 to 0.9 reduces mean wall latency from 23.93~s to 14.43~s, and EfficientNet-B3 backbones run faster on average than ResNet-50 backbones.}
\label{tab:post-dp}
\setlength{\tabcolsep}{3pt} %
\resizebox{\textwidth}{!}{%
\begin{tabular}{llccccccccccccccccc}
\toprule
\multirow{2}{*}{\textbf{Model}} & \multirow{2}{*}{\textbf{Backbone}}
& \multicolumn{3}{c}{\textbf{Pruning Ratio = 0.5}}
& \multicolumn{3}{c}{\textbf{Pruning Ratio = 0.6}}
& \multicolumn{3}{c}{\textbf{Pruning Ratio = 0.7}}
& \multicolumn{3}{c}{\textbf{Pruning Ratio = 0.8}}
& \multicolumn{3}{c}{\textbf{Pruning Ratio = 0.9}} \\
\cmidrule(lr){3-5} \cmidrule(lr){6-8} \cmidrule(lr){9-11} \cmidrule(lr){12-14} \cmidrule(lr){15-17}
& & \multicolumn{2}{c}{\textbf{Latency (s)}} & \textbf{Temp}
  & \multicolumn{2}{c}{\textbf{Latency (s)}} & \textbf{Temp}
  & \multicolumn{2}{c}{\textbf{Latency (s)}} & \textbf{Temp}
  & \multicolumn{2}{c}{\textbf{Latency (s)}} & \textbf{Temp}
  & \multicolumn{2}{c}{\textbf{Latency (s)}} & \textbf{Temp} \\
\cmidrule(lr){3-4} \cmidrule(lr){6-7} \cmidrule(lr){9-10} \cmidrule(lr){12-13} \cmidrule(lr){15-16}
& & \textbf{Wall} & \textbf{CPU} & (°C)
  & \textbf{Wall} & \textbf{CPU} & (°C)
  & \textbf{Wall} & \textbf{CPU} & (°C)
  & \textbf{Wall} & \textbf{CPU} & (°C)
  & \textbf{Wall} & \textbf{CPU} & (°C) \\
\midrule
\multirow{2}{*}{DeepLabV3+}
& ResNet-50        & 25.9 & 12.4 & 52.1  & 24.2 & 11.9 & 51.6  & 21.5 & 11.14 & 51.6  & 19.5 & 10.6 & 51.6  & 16.7 & 9.9 & 51.1 \\
& EfficientNet-B3  & 23.5 & 11.4 & 52.5  & 19.7 & 10.4 & 51.6  & 15.7 & 9.4 & 53.5  & 13.5 & 8.7 & 52.1  & 11.5 & 8.3 & 51.6 \\
\midrule
\multirow{2}{*}{U-Net}
& ResNet-50        & 28.9 & 13.1 & 52.5  & 23.4 & 9.1 & 53  & 24.7 & 12.1 & 53  & 22.1 & 11.3 & 53  & 19.9 & 10.7 & 52.5 \\
& EfficientNet-B3  & 20.3 & 8.1 & 51.6  & 17.5 & 7.5 & 51.6  & 14.7 & 6.3 & 51.1  & 15.8 & 8.6 & 50.1  & 11.8 & 5.9 & 49.1 \\
\midrule
\multirow{2}{*}{FPN}
& ResNet-50        & 22.9 & 11.5 & 51.1  & 20.5 & 10.9 & 52.5  & 18.6 & 10.3 & 52.1  & 17.1 & 10.0 & 50.6  & 14.9 & 9.4 & 51.1 \\
& EfficientNet-B3  & 22.1 & 11.0 & 52.5  & 18.7 & 10.2 & 52.1  & 15.5 & 9.4 & 52.5  & 13.6 & 8.9 & 52.1  & 11.8 & 8.4 & 52.1 \\
\bottomrule
\end{tabular}
}
\end{table*}

\subsubsection{Quantization Results}
The quantization results show that the optimal quantization level is model-dependent rather than strictly proportional to bit-width. For Qwen3.5 0.8B, Q5\_K\_M provides the best overall trade-off, achieving the highest quality mean of 59.68 with a GGUF size of 551.22 MB. It also gives the strongest SQuAD scores among Qwen configurations, with 93.85 F1 and 92 EM, while maintaining 75\% BoolQ accuracy. Although Q4\_K\_M further reduces the model size to 507.85 MB, its quality mean drops to 54.92, indicating that aggressive quantization begins to degrade Qwen's QA performance.

For TinyLlama 1.1B Chat and Gemma 3 1B IT, Q4\_K\_M performs unexpectedly well. TinyLlama obtains its best quality mean under Q4\_K\_M, reaching 46.44 with a reduced size of 637.81 MB, outperforming its Q8\_0, Q6\_K, and Q5\_K\_M variants. Similarly, Gemma reaches its best baseline quality mean of 47.35 under Q4\_K\_M with a size of 768.72 MB. In these fixed-prompt QA runs, quality is therefore non-monotone in bit width: for TinyLlama and Gemma the smallest artifact also carries the highest aggregate score.

For deployment, Q4\_K\_M is the memory-first choice, whereas Q5\_K\_M preserves the strongest Qwen QA quality.

This non-monotonic behavior also separates our observations from the post-training quantization literature: GPTQ \cite{frantar2022gptq}, SmoothQuant \cite{xiao2022smoothquant}, AWQ \cite{lin2023awq}, and QLoRA \cite{dettmers2023qlora} are framed around preserving accuracy while reducing memory, asking how close a low-bit model can stay to higher precision. In our small-model QA runs the lower-bit artifact is at times not merely close but better on specific extractive metrics, with the prompt template and the sample identifiers held fixed across bit widths, indicating that quantization changes generation behavior rather than merely degrading numerical fidelity.

\subsection{Image Model Results}

Image segmentation results were taken from our recently published work \cite{Our_CCNC_2026work}. That work proposed SPICE, an end-to-end model-selection, optimization, and deployment pipeline for real-world edge inference on spotted lanternfly images. The task was to detect spotted lanternflies in field images and classify their life-cycle stages, including adult, egg masses, instar nymph stages 1--3, and instar nymph stage 4. In that work, DeepLabV3+, U-Net, and FPN were evaluated with ResNet-50 and EfficientNet-B3 backbones to establish strong full-size baselines before compression. The broader deployment objective was to make these models feasible on extremely resource-constrained platforms, such as drone-mounted edge devices, so that spotted lanternflies can be automatically detected while the platform hovers over agricultural fields.

\subsubsection{Baseline Results}

Table~\ref{tab:1} reports the full-size baseline performance of six segmentation models. Among the full models, FPN with EfficientNet-B3 achieved the highest mIoU of 0.6728 and the highest mean accuracy of 75.59\%, despite using only 12.476M parameters, 16.33G MACs, and 47.19 MB of storage. DeepLabV3+ with EfficientNet-B3 was similarly compact, requiring 11.681M parameters, 15.02G MACs, and 44.22 MB, while maintaining an mIoU of 0.6611 and accuracy of 74.13\%. The heaviest baseline was U-Net with ResNet-50, requiring 32.521M parameters, 54.02G MACs, and 130.99 MB, but it achieved only 0.5612 mIoU and 61.91\% accuracy. This shows that the largest model was not the most accurate: compared with U-Net--ResNet-50, FPN--EfficientNet-B3 improved mIoU by 0.1116 and accuracy by 13.68 percentage points while using 61.64\% fewer parameters, 69.77\% fewer MACs, and 63.97\% less storage. Averaged across all full-size models, the baseline mIoU was 0.6385 and the accuracy was 71.63\%, with model sizes ranging from 44.22 MB to 130.99 MB.

\subsubsection{Pruning Results}

TaLK-Structure Pruning combines first-order Taylor sensitivity for inter-layer pruning-quota allocation, L2-norm-based channel selection for intra-layer pruning, and knowledge-distillation-assisted fine-tuning for accuracy recovery \cite{Our_CCNC_2026work}. Unlike unstructured pruning, which only inserts sparse masks and therefore may not reduce runtime or memory on general-purpose edge hardware, structured pruning removes complete channels or filters and directly reduces parameter count, MACs, and model size. As shown in Table~\ref{tab:1}, at pruning ratio $S=0.5$, structured pruning reduced the average parameter count by 61.55\%, MACs by 43.14\%, and model size by 61.98\% across the six models. U-Net--ResNet-50 obtained the largest storage reduction, from 130.99 MB to 26.69 MB, corresponding to a 79.62\% decrease, while its mIoU decreased only from 0.5612 to 0.5517 and its accuracy increased slightly from 61.91\% to 62.86\%. DeepLabV3+--ResNet-50 offered the strongest pruned accuracy--compression trade-off, reducing parameters from 26.678M to 7.943M, MACs from 46.29G to 20.78G, and size from 106.71 MB to 30.30 MB, while retaining 0.6258 mIoU and 70.97\% accuracy. In contrast, EfficientNet-B3-backed models were more compression-sensitive: U-Net--EfficientNet-B3 dropped from 0.6636 to 0.4467 mIoU and from 74.80\% to 54.76\% accuracy, indicating that compact backbones can be less tolerant to aggressive structural channel removal.

\subsubsection{Quantization Results}

Table~\ref{tab:1} shows that default quantization produced almost no architectural compression: parameter count and MACs remained unchanged for all six models, while model size decreased only marginally. The average storage reduction was only 0.63\%, with the largest reduction observed for DeepLabV3+--EfficientNet-B3, from 44.22 MB to 43.02 MB, i.e., only 2.71\%. The underlying workflow in our implementation --- PyTorch FX graph-mode quantization with the stock \texttt{fbgemm} configuration, coupled with a brief quantization-aware fine-tuning stage and serialized via TorchScript --- therefore did not yield a compressed deployable artifact for these convolution- and upsampling-dominated architectures, which are not fully quantizable under PyTorch's current backends~\cite{Our_CCNC_2026work}. Parameter and MAC counts describe the topology rather than the numerical format, so they are unaffected by construction; the storage figures are what reveal that no effective compression took place. As a result, quantized mIoU remained nearly identical to the full model, with an average change of only $-0.0024$. Some accuracy values increased after quantization, such as FPN--ResNet-50 from 69.48\% to 71.48\% and U-Net--ResNet-50 from 61.91\% to 64.31\%, but because the workflow includes that brief quantization-aware fine-tuning stage, these increases do not isolate an effect of the numerical format.

\subsubsection{Post-deployment Statistics}

Table~\ref{tab:post-dp} reports Raspberry Pi deployment statistics for structured-pruned models across pruning ratios from 0.5 to 0.9. Across all 30 deployment configurations, the average wall latency was 18.88 s and the average CPU latency was 9.89 s, with temperatures remaining stable at 51.85$^\circ$C on average. Increasing the pruning ratio from 0.5 to 0.9 reduced the mean wall latency from 23.93 s to 14.43 s, corresponding to a 39.69\% reduction, while mean CPU latency decreased from 11.25 s to 8.77 s, corresponding to a 22.07\% reduction. The fastest wall-clock inference was achieved by DeepLabV3+--EfficientNet-B3 at pruning ratio 0.9, with 11.5 s wall latency and 8.3 s CPU latency, whereas the slowest configuration was U-Net--ResNet-50 at pruning ratio 0.5, with 28.9 s wall latency and 13.1 s CPU latency. EfficientNet-B3 backbones were faster on average than ResNet-50 backbones, with mean wall latency of 16.38 s compared with 21.39 s. However, the accuracy results in Table~\ref{tab:1} show that this latency advantage must be balanced against pruning sensitivity, since ResNet-50-backed models, especially DeepLabV3+--ResNet-50, retained stronger post-pruning task performance.

\begin{table}[t]
\centering
\caption{BoolQ prediction skew under compression. All runs use the
\texttt{Q4\_K\_M} GGUF format on GPU over the same 100-sample subset (gold prior: 70\% yes, 30\% no). Strict accuracy is misled in opposite directions by the two model families, while balanced accuracy exposes that both collapse toward a constant prior.}
\label{tab:boolq_prior_collapse}%
\footnotesize
\setlength{\tabcolsep}{3pt}
\renewcommand{\arraystretch}{1.1}
\begin{tabular*}{\columnwidth}{@{\extracolsep{\fill}}
l l c c c c c @{}}
\toprule
\onehead{Model} &
\onehead{Config} &
\twohead{Parse}{(\%)} &
\twohead{Yes}{(\%)} &
\twohead{No}{(\%)} &
\twohead{Strict}{Acc.} &
\twohead{Bal.}{Acc.} \\
\midrule
\multicolumn{7}{@{}l}{\textbf{Qwen2.5 1.5B Instruct}} \\
\midrule
& Baseline        & 100 & 75 & 25 & 75 & 67.9 \\
& Pruned $3\%$    & 78  & 78 & 0  & 53 & 37.9 \\
& {+}LoRA $1\%$   & 100 & 97 & 3  & 71 & 52.6 \\
\midrule
\multicolumn{7}{@{}l}{\textbf{Qwen3.5 0.8B}} \\
\midrule
& Baseline        & 100 & 47 & 53 & 73 & 78.8 \\
& Pruned $3\%$    & 100 & 6  & 94 & 36 & 54.3 \\
& Pruned $5\%$    & 100 & 1  & 99 & 31 & 50.7 \\
\bottomrule
\end{tabular*}
\end{table}

\begin{figure*}[!t]
\centering
\includegraphics[scale=0.85]{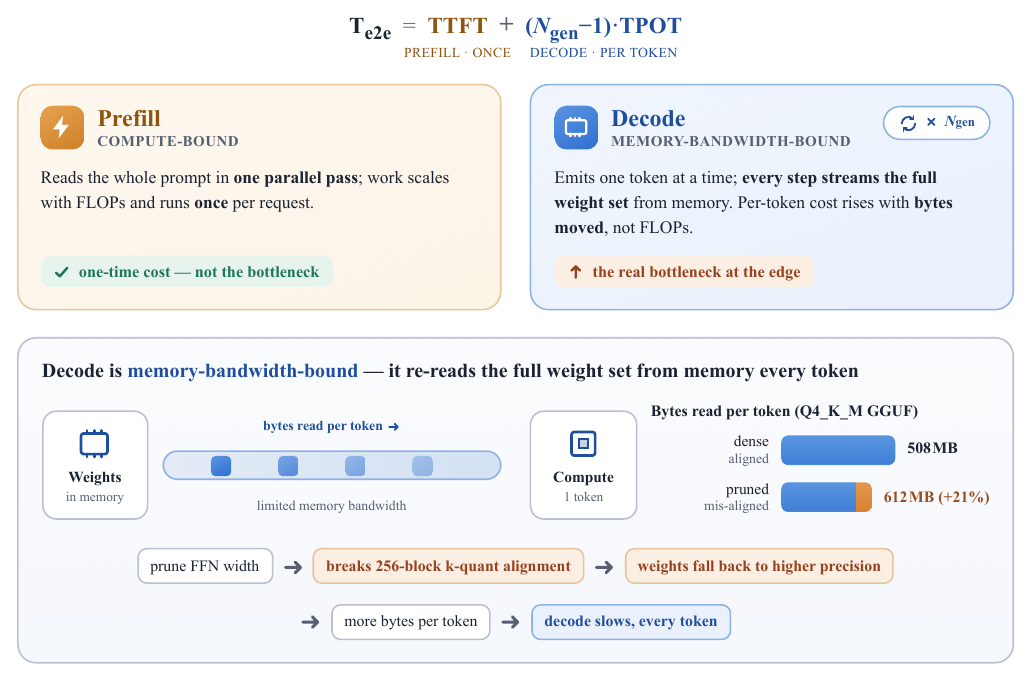}
\caption{A prefill/decode view of why structured pruning can slow edge
inference. End-to-end latency splits into a one-off compute-bound prefill and a
per-token memory-bandwidth-bound decode; because $k$-quant fallback inflates the
deployable weight bytes (508 to 612~MB), decode re-reads a larger weight set on
each of the $N_{\text{gen}}$ output tokens. Measured phase changes appear in Table~\ref{tab:phase_split_measured}.}
\label{fig:phase_split}
\end{figure*}

\section{Observations}

\subsection{Pruning-Induced Forgetfulness}

At approximately $1\%$ MLP-channel pruning, SQuAD F1 falls by 17.8--64.5 points across all three evaluated LLM families in the internally controlled GPU and CPU comparisons; changes on BoolQ and NQ remain model- and platform-specific. Pruning at this ratio therefore removes structures that carry task-specific performance, an effect we refer to as pruning-induced forgetfulness. To analyze this behavior, we model a neural network as a directed weighted computational graph --- hidden units or intermediate representations as nodes, parameterized transformations as weighted edges --- following prior pruning studies that formulate networks as graph-structured computational systems~\cite{fang2023depgraph} \cite{fang2024isomorphic}. Pruning then removes selected edges, channels, or substructures from this graph, and its effect is a distortion of the signal flow from input to output. The same view underlies sensitivity-based criteria such as Taylor and empirical Fisher scores, which use gradient information to estimate how such local perturbations affect the network output.

Let $G=(V,E,W)$ be a directed weighted computational graph representing a neural network. Each prunable linear layer can be viewed as a bipartite subgraph between two consecutive hidden representations. For a prunable linear operator in layer $k$, let
\begin{itemize}
    \item $\mathbf{H}^{(k-1)}\in\mathbb{R}^{T\times d_{\mathrm{in}}}$ denote the
    input activation matrix over $T$ tokens, computed from the dense model for a
    fixed input sample;
    \item $T$ denote the number of tokens and $d_{\mathrm{in}}$ denote the input
    hidden dimension;
    \item $\mathbf{W}^{(k)}\in\mathbb{R}^{d_{\mathrm{out}}\times d_{\mathrm{in}}}$
    denote the corresponding weight matrix;
    \item $\mathbf{b}^{(k)}\in\mathbb{R}^{d_{\mathrm{out}}}$ denote the bias
    vector, broadcast over the token dimension;
    \item $d_{\mathrm{out}}$ denote the output hidden dimension of layer $k$.
\end{itemize}

For a fixed input sample, the dense pre-activation flow is
\begin{equation}
    \mathbf{Z}^{(k)}
    =
    \mathbf{H}^{(k-1)}
    \left(\mathbf{W}^{(k)}\right)^{\top}
    +
    \mathbf{1}_{T}
    \left(\mathbf{b}^{(k)}\right)^{\top},
    \qquad
    \mathbf{Z}^{(k)}\in\mathbb{R}^{T\times d_{\mathrm{out}}},
\end{equation}
where $\mathbf{1}_{T}\in\mathbb{R}^{T}$ is an all-one vector used to broadcast
the bias across all tokens.

Pruning is modeled by a binary mask
$\mathbf{M}^{(k)}\in\{0,1\}^{d_{\mathrm{out}}\times d_{\mathrm{in}}}$, giving
the pruned weight matrix
\begin{equation}
    \widetilde{\mathbf{W}}^{(k)}
    =
    \mathbf{M}^{(k)}\odot \mathbf{W}^{(k)} .
\end{equation}
Here, $M_{ij}^{(k)}=1$ keeps the corresponding weight, whereas
$M_{ij}^{(k)}=0$ removes it. We define the complement pruning mask as
\begin{equation}
    \overline{\mathbf{M}}^{(k)}
    =
    \mathbf{1}_{d_{\mathrm{out}}\times d_{\mathrm{in}}}
    -
    \mathbf{M}^{(k)},
    \qquad
    \overline{\mathbf{M}}^{(k)}
    \in
    \{0,1\}^{d_{\mathrm{out}}\times d_{\mathrm{in}}},
\end{equation}
where $\overline{\mathbf{M}}^{(k)}$ explicitly identifies the removed weights.

Under a fixed-activation approximation, the activations
$\mathbf{H}^{(k-1)}$ are taken from the dense model and used as a first-order proxy. This approximation measures the immediate local distortion caused by pruning, while cascading changes from earlier layers are not explicitly modeled. The pruning-induced pre-activation flow deficit at layer $k$ is therefore
\begin{equation}
    \Delta \mathbf{Z}^{(k)}
    =
    \mathbf{Z}^{(k)}-\widetilde{\mathbf{Z}}^{(k)}
    =
    \mathbf{H}^{(k-1)}
    \left(
    \overline{\mathbf{M}}^{(k)}
    \odot
    \mathbf{W}^{(k)}
    \right)^{\top}.
\end{equation}

Since $\Delta \mathbf{Z}^{(k)}\in\mathbb{R}^{T\times d_{\mathrm{out}}}$, the quantity $T d_{\mathrm{out}}$ denotes the total number of scalar pre-activation values in layer $k$. We therefore normalize the squared Frobenius norm by
$T d_{\mathrm{out}}$ and define the layer-level pruning distortion as
\begin{equation}
    \mathcal{D}_{\mathrm{layer}}^{(k)}
    =
    \mathbb{E}_{\mathbf{x}\sim\mathcal{D}}
    \left[
    \frac{1}{T d_{\mathrm{out}}}
    \left\|
    \Delta \mathbf{Z}^{(k)}
    \right\|_{F}^{2}
    \right],
\end{equation}
where the expectation is taken over the data distribution $\mathcal{D}$. Thus, $\mathbf{H}^{(k-1)}$ in the previous equation denotes the activation matrix for
one input sample, while $\mathcal{D}_{\mathrm{layer}}^{(k)}$ averages this
distortion across samples.

For a network with $K$ prunable layers, the total pruning distortion is written as
\begin{equation}
    \mathcal{D}_{\mathrm{prune}}
    =
    \sum_{k=1}^{K}
    \lambda_k
    \mathcal{D}_{\mathrm{layer}}^{(k)},
    \qquad
    \lambda_k\geq 0 .
\end{equation}
Here, $\lambda_k$ is a layer-wise weighting factor. If all layers are treated
equally, one may set $\lambda_k=1$ for all $k$; otherwise, $\lambda_k$ can be
used to emphasize more sensitive layers.

This formulation shows that pruning acts as a multiplicative masking operation on the weights through $\mathbf{M}^{(k)}\odot\mathbf{W}^{(k)}$, while the total distortion accumulates as a weighted additive sum across layers. As the pruning ratio increases, more entries become active in the complement mask $\overline{\mathbf{M}}^{(k)}$, which generally increases the local flow distortion. This motivates fine-tuning after pruning, so that the remaining weights can adapt and partially recover the disrupted signal flow.

We instantiate the measure for every pruned configuration in Table~\ref{tab:gpu_cpu_rpi_pruned_llm_results}, computing $\mathcal{D}_{\mathrm{prune}}$ with $\lambda_k=1$ over all gate, up, and down MLP projections from dense-model activations on 100 SQuAD prompts; the archived pruned models retain their widths but not their masks, so removed-channel identities are reconstructed by re-running the deterministic group-magnitude pruning rule against the saved widths. Table~\ref{tab:distortion_measured} reports the values. Within a family, $\mathcal{D}_{\mathrm{prune}}$ orders the damage where the damage is graded: for Qwen3.5 its rank matches the measured SQuAD F1 drop exactly on both GPU and CPU (Spearman $\rho=1.0$; Pearson $r=0.97$ and $0.96$; with four ratios per family, the exact permutation $p$ floors at $0.083$), and the supplemental Qwen2.5 trajectory is likewise rank-perfect ($\rho=1.0$). TinyLlama is only moderately aligned ($\rho=0.8$): its $3\%$ configuration damages less than its $1\%$. Gemma sits at the task floor at every ratio (F1 drops of $63.2$--$65.0$ points), leaving no ordering to test. In every measured trajectory, $\mathcal{D}_{\mathrm{prune}}$ grows with the pruning ratio, so these four-point trajectories certify consistency with the measured damage rather than predictive power beyond the ratio itself. Raw magnitudes also do not transfer across architectures: Qwen2.5 at $1\%$ pruning registers $\mathcal{D}_{\mathrm{prune}}=2.77$, above Gemma at $7\%$ ($2.72$), yet loses $3.2$ F1 points against Gemma's $64.5$. The layer decomposition explains why: the projection carrying the distortion differs by architecture --- at $1\%$ pruning the gate projection holds $77\%$ of Qwen3.5's total against $1\%$ on its down projection, whereas Qwen2.5 concentrates $67\%$ on the down projection --- so an unnormalized sum over projections with architecture-specific roles cannot share a scale across families. $\mathcal{D}_{\mathrm{prune}}$ is therefore a within-family severity diagnostic, not a cross-model damage predictor.

\begin{table}[!t]
\centering
\caption{Fixed-activation pruning distortion $\mathcal{D}_{\mathrm{prune}}$ computed on 100 SQuAD prompts with $\lambda_k=1$. Raw values are comparable only within a row; Spearman $\rho$ relates each family's four nested pruning configurations to its measured SQuAD F1 drop on GPU\,/\,CPU. Gemma is at the task floor at every ratio, so no coefficient is reported.}
\label{tab:distortion_measured}
\small
\begin{tabular}{lccccc}
\toprule
\textbf{Model} & \textbf{1\%} & \textbf{3\%} & \textbf{5\%} & \textbf{7\%} & \textbf{Spearman $\rho$} \\
\midrule
Gemma-1B  & 0.719 & 1.482 & 2.252 & 2.719 & task floor \\
Qwen3.5   & 0.189 & 0.445 & 0.638 & 0.817 & 1.00 / 1.00 \\
TinyLlama & 0.107 & 0.261 & 0.383 & 0.491 & 0.80 / 0.80 \\
\bottomrule
\end{tabular}
\end{table}

\subsection{BoolQ Prior Collapse} A recurring evaluation hazard on BoolQ~\cite{clark2019boolq} is that a compressed model can stay fully parseable while collapsing into a degenerate yes/no prior, so that strict accuracy reports comprehension that is no longer there. Because the gold labels are imbalanced---roughly $70\%$ yes in our sampled setting---this collapse misleads strict accuracy in either direction, and the direction depends on the model rather than on the amount of damage.
Table~\ref{tab:boolq_prior_collapse} reports this skew for two Qwen models under matched \texttt{Q4\_K\_M} GPU evaluation. The analysis pairs Qwen2.5 1.5B Instruct with the Qwen3.5 0.8B used throughout the rest of this study: the two collapse toward opposite classes, so the pair separates the pathology from any single model's answer bias.

The two families collapse toward opposite constants. Qwen3.5 0.8B at $5\%$ pruning answers no almost unconditionally (a single yes prediction in $100$ samples, with parsing still valid), so its strict accuracy falls to the always-no floor near $31\%$ and the damage is obvious. Qwen2.5 1.5B at $3\%$ pruning shows the mirrored failure: it answers yes on every parsed sample, and because the prior is yes-heavy it still retains $67.9\%$ \emph{parsed} accuracy while discriminating nothing. Read through strict accuracy alone, the ranking of the two collapsed configurations inverts: Qwen2.5 at $3\%$ pruning scores $53$ against Qwen3.5's $31$ at $5\%$, while balanced accuracy places them the other way round, at $37.9$ and $50.7$.

LoRA recovery can sharpen rather than cure this illusion. Qwen2.5 1.5B with  LoRA at $1\%$ pruning predicts yes on $97$ of $100$ samples and keeps a high strict accuracy of $71\%$, yet its balanced accuracy is only $52.6\%$---barely
above chance---which is prior recovery, not comprehension recovery. The
balanced-accuracy column makes the shared pathology explicit: no collapsed
configuration rises above $54.3\%$ regardless of how healthy its strict score
appears. BoolQ results under compression should therefore be read together with
parse rate, prediction skew, and balanced accuracy rather than strict accuracy
in isolation.

\subsection{A Prefill/Decode View of the Pruning Slowdown}
\label{subsec:prefill_decode}

End-to-end latency is clearest when split into the two phases of autoregressive inference. Prefill consumes the whole prompt in a single parallel pass and is compute-bound. Decode then emits one token at a time, and because every step must re-read the full weight set from memory, its cost is governed by memory bandwidth rather than arithmetic~\cite{pope2022efficiently,patel2023splitwise}. A request therefore costs
\begin{equation}
T_{\text{e2e}} = \underbrace{\text{TTFT}}_{\text{prefill}}
  + \underbrace{(N_{\text{gen}}-1)\,\text{TPOT}}_{\text{decode}},
\label{eq:phase_decomposition}
\end{equation}
where TTFT (time to first token) measures the one-off prefill and TPOT (time per
output token) measures the per-token decode cost.

This split explains why structured pruning slowed inference on every internally controlled CPU comparison despite removing parameters. Pruning removes parameters, but once the pruned width breaks the $k$-quant block alignment of \eqref{eq:kquant_alignment}, the affected tensors fall back to a higher-precision encoding and the deployable weights grow rather than shrink --- in our screening the Q4\_K\_M weights rise from 508 to 612~MB ($+21\%$).
Every one of the $N_{\text{gen}}$ output tokens must re-read this larger weight
set, so the memory-bandwidth-bound decode phase pays the inflated cost on each
step, whereas the compute-bound prefill pays it only once. The slowdown therefore
accumulates in decode, exactly where edge inference is already most constrained
(Fig.~\ref{fig:phase_split}).

The quantization-sweep serving logs measure this split directly. \texttt{llama.cpp} logs a prompt-evaluation duration and a generation duration per request; we take the former as the prefill proxy and, over requests generating more than one token, total generation time divided by generated intervals as the token-weighted per-token decode cost. Every logged sweep run executes with all layers GPU-offloaded, and each configuration was served twice at 300 requests per run (18{,}000 requests in total), so the two recorded run sets bound run-to-run variation. Across all 18 matched baseline-versus-pruned pairs --- three families at \texttt{Q4\_K\_M}, \texttt{Q5\_K\_M}, and \texttt{Q6\_K}, both run sets --- the generation tail grows faster than prompt evaluation, and token-weighted per-token time rises in all 18. Table~\ref{tab:phase_split_measured} locates the slowdown: Gemma's is almost entirely verbosity ($+169\%$ generated tokens against $+6\%$ per token), TinyLlama's is mostly per-token cost ($+25$--$26\%$ per token against $+11\%$ tokens), and Qwen3.5 mixes the two. The generation caps (48 SQuAD, 16 BoolQ, 48 NQ tokens) censor the longest pruned outputs, so these token-count increases are lower bounds. Prompt evaluation moves by at most $+16\%$: the higher-precision fallback tensors raise its cost slightly rather than lowering it.

\begin{table}[!t]
\centering
\caption{Measured phase changes after unadapted DG-MLP pruning, relative to the dense baseline in the quantization-sweep serving logs. Each entry spans the two recorded GPU-offloaded run sets, averaged over \texttt{Q4\_K\_M}, \texttt{Q5\_K\_M}, and \texttt{Q6\_K}; generated-token counts coincide across the run sets. In every family the generation tail grows several times faster than prompt evaluation.}
\label{tab:phase_split_measured}
\small
\setlength{\tabcolsep}{3.5pt}
\begin{tabular}{lccccc}
\toprule
\textbf{Model} & \textbf{Prompt eval} & \textbf{Per token} & \textbf{Tokens} & \textbf{Gen.\ tail} & \textbf{End-to-end} \\
\midrule
Gemma-1B (1\%)  & $+3.2$--$3.3\%$  & $+6.1$--$6.2\%$  & $+168.5\%$ & $+202$--$203\%$ & $+147$--$149\%$ \\
Qwen3.5 (1\%)   & $+2.2$--$4.2\%$  & $+3.0$--$7.0\%$  & $+40.8\%$  & $+54$--$60\%$   & $+29$--$30\%$ \\
TinyLlama (3\%) & $+13.4$--$16.0\%$ & $+24.7$--$25.7\%$ & $+10.9\%$  & $+40$--$41\%$   & $+36$--$37\%$ \\
\bottomrule
\end{tabular}
\end{table}

\begin{table*}[!t]
\centering
\caption{Use-case-specific deployment guidance for LLM downstream tasks.}
\label{tab:deployment_guidance_llm}
\scriptsize
\renewcommand{\arraystretch}{1.18}
\setlength{\tabcolsep}{3pt}
\begin{tabularx}{\textwidth}{
>{\raggedright\arraybackslash}p{0.12\textwidth}
>{\raggedright\arraybackslash}p{0.24\textwidth}
>{\raggedright\arraybackslash}p{0.17\textwidth}
>{\raggedright\arraybackslash}X
}
\toprule
\textbf{Downstream Task} &
\textbf{Representative Edge Use Cases} &
\textbf{Primary Constraint} &
\textbf{Evidence and Deployment Recommendation} \\
\midrule

SQuAD / Extractive QA &
On-device WhatsApp or Telegram chatbot, voice-based question answering, educational tutor, field-manual assistant, offline maintenance guide, local document QA, disaster-response information assistant. \cite{zhang2024edgeshard} \cite{luo2025toward} &
Latency is the common constraint for all LLM tasks; for SQuAD, answer precision and short-form output discipline are also critical. &
Table~\ref{tab:gpu_cpu_rpi_pruned_llm_results} shows that Qwen3.5~0.8B provides the strongest unpruned SQuAD performance, achieving 84.08 F1 on GPU, 86.30 F1 on CPU, and 87.93 F1 on Raspberry Pi. However, DepGraph-MLP pruning severely reduces extractive QA quality: Qwen drops from 84.08 to 21.63 GPU SQuAD F1 at 7.03\% pruning, while Raspberry Pi latency can increase up to 25.84~s/sample at 3.01\% pruning, measured across campaigns whose prompt templates are not controlled. Therefore, for extractive QA, the preferred deployment choice is the unpruned or quantized Qwen model rather than an aggressively pruned variant. Table~\ref{tab:quant_prune_lora_results} further shows that Q5\_K\_M is the best Qwen quantization point, giving 93.85 SQuAD F1, 92 EM, and 551.22~MB GGUF size, while Q4\_K\_M reduces the size to 507.85~MB but lowers the quality mean. \\

\midrule

Natural Questions / Open-domain QA &
Offline search assistant, local RAG assistant, field-survey knowledge lookup, emergency-response query system, personal document assistant, agricultural advisory assistant. &
Accuracy and answer reliability dominate; latency remains important, but the model must preserve factual recall and avoid answer drift. &
Natural Questions is more fragile than SQuAD in our compressed LLM experiments. In Table~\ref{tab:gpu_cpu_rpi_pruned_llm_results}, even the best unpruned Qwen baseline reports only 7.69/2 NQ F1/EM on GPU, 8.18/2 on CPU, and 10.22/4 on Raspberry Pi. LoRA recovery is not uniformly restorative: for Qwen at 1\% pruning, LoRA improves BoolQ but reduces SQuAD F1 and NQ F1; at 5\% pruning, LoRA improves SQuAD and BoolQ but NQ remains degraded. Therefore, open-domain QA should use the strongest unpruned or lightly quantized baseline, preferably Qwen with Q5\_K\_M when memory is constrained. Pruning should be avoided unless the post-pruning model is re-evaluated directly on NQ-style queries. \\

\midrule

BoolQ / Binary QA &
Yes/no command validation, smart-home confirmation, robotic decision checks, clinical or agricultural checklist assistant, local policy-compliance assistant, FAQ triage. \cite{hou2025applied} \cite{raza2025industrial}&
Latency and parseability are necessary, but balanced accuracy, prediction skew, and yes/no prior collapse must also be checked. &
BoolQ accuracy alone is insufficient under compression. Table~\ref{tab:gpu_cpu_rpi_pruned_llm_results} shows that Qwen baseline achieves 73\% BoolQ accuracy on GPU, 70\% on CPU, and 66.67\% parsed BoolQ accuracy on Raspberry Pi. However, compressed models may remain parseable while collapsing into degenerate yes/no priors. For example, Qwen3.5 at 5\% pruning almost always predicts ``no,'' while another pruned model configuration can preserve a high strict score by predicting the majority ``yes'' class. Therefore, BoolQ deployments should not select models using strict accuracy alone. They should report parsed accuracy, parse rate, prediction skew, and balanced accuracy; LoRA recovery should also be validated per task because it can improve BoolQ while degrading SQuAD or NQ. \\

\midrule

Chat / Free-form Generation &
Local conversational agent, voice assistant, lightweight customer-support bot, tutoring assistant, text rewriting assistant, on-device command interface, privacy-preserving personal assistant~\cite{yang2024talk2care} \cite{molina2024leveraging} \cite{li2024personal} &
Latency, tokens per second, response length control, and output-format stability. &
Figure~\ref{fig:tps} and Table~\ref{tab:gpu_cpu_rpi_pruned_llm_results} indicate that pruning does not guarantee faster generation. Although pruning removes MLP channels, it can make outputs longer and less format-compliant, increasing per-sample latency. For Qwen3.5, Raspberry Pi latency increases from 7.53~s/sample in the baseline to 25.84~s/sample at 3.01\% pruning, measured across campaigns whose prompt templates are not controlled; the controlled CPU comparison moves in the same direction. Therefore, free-form chat deployments should prioritize quantization and decoding control over structural pruning. A practical policy is to use Q4\_K\_M when memory is the dominant constraint and Q5\_K\_M when answer quality is more important, while enforcing a maximum output length to prevent latency blow-up. This guidance extrapolates from the measured generation-latency behavior of the QA workloads; chat quality itself was not separately benchmarked. \\

\bottomrule
\end{tabularx}
\end{table*}

\begin{table*}[!t]
\centering
\caption{Use-case-specific deployment guidance for image-model downstream tasks.}
\label{tab:deployment_guidance_image}
\scriptsize
\renewcommand{\arraystretch}{1.18}
\setlength{\tabcolsep}{3pt}
\begin{tabularx}{\textwidth}{
>{\raggedright\arraybackslash}p{0.12\textwidth}
>{\raggedright\arraybackslash}p{0.24\textwidth}
>{\raggedright\arraybackslash}p{0.17\textwidth}
>{\raggedright\arraybackslash}X
}
\toprule
\textbf{Downstream Task} &
\textbf{Representative Edge Use Cases} &
\textbf{Primary Constraint} &
\textbf{Evidence and Deployment Recommendation} \\
\midrule

Semantic Segmentation &
Drone-based pest monitoring, crop-field surveillance, biological image segmentation, flood-scene parsing, robot navigation, field inspection, on-device ecological monitoring. \cite{yang2023device}&
mIoU, wall latency, CPU latency, memory footprint, and thermal stability. &
Table~\ref{tab:1} shows that the best full-size segmentation model is FPN--EfficientNet-B3, with 0.6728 mIoU, 75.59\% mean accuracy, 12.476M parameters, 16.33G MACs, and 47.19~MB size. However, under structured pruning at $S=0.5$, DeepLabV3+--ResNet-50 gives the best accuracy--compression trade-off, retaining 0.6258 mIoU and 70.97\% accuracy while reducing size from 106.71~MB to 30.30~MB. Table~\ref{tab:post-dp} further shows that increasing pruning from 0.5 to 0.9 reduces average wall latency from 23.93~s to 14.43~s. Therefore, when accuracy dominates, the full FPN--EfficientNet-B3 model is preferred; when edge feasibility dominates, DeepLabV3+--ResNet-50 at moderate pruning is the safer choice; when latency dominates, high-pruning EfficientNet-backed models should be considered only after verifying mIoU degradation. \\

\midrule

Image Classification / Stage Recognition &
Spotted-lanternfly life-stage recognition, weed classification, leaf-disease screening, visual inspection, local camera-based monitoring, binary or multi-class field recognition. \cite{zhang2026real} &
Classification accuracy, memory footprint, and inference latency. &
Table~\ref{tab:1} reports mean accuracy jointly with segmentation mIoU. FPN--EfficientNet-B3 gives the highest full-size mean accuracy of 75.59\%, while the quantized version gives 75.77\% but with almost no real reduction in parameters or MACs. Structured pruning gives much stronger compression: DeepLabV3+--ResNet-50 reduces size from 106.71~MB to 30.30~MB while retaining 70.97\% mean accuracy. For resource-constrained image-classification or post-segmentation stage-recognition tasks, structured pruning is therefore more useful than default quantization when the target is actual edge deployment. However, if the deployment device has sufficient memory and accuracy is the dominant criterion, the full FPN--EfficientNet-B3 model remains the preferred candidate. This guidance extrapolates from segmentation mean accuracy; image classification was not separately benchmarked. \\

\bottomrule
\end{tabularx}
\end{table*}

\section{Deployment Recommendations}
The effectiveness of a compression method depends not only on the compression ratio, but also on the downstream task, the target hardware, the deployment format, and the acceptable trade-off between accuracy, latency, memory footprint, and energy consumption. Therefore, deployment decisions should not be made from model size alone. Instead, each candidate model should be evaluated against task-specific constraints, such as extractive-answer quality for question answering, parse reliability for binary reasoning, mIoU for segmentation, classification accuracy for image recognition, and wall-clock latency for real-time edge inference.

\subsection{Use-case Specific Selection}
\label{subsec:usecase_specific_selection}

In our runs the same compression step moves tasks in opposite directions: LoRA recovery at 1\% pruning raises Qwen3.5's BoolQ accuracy from 57 to 72 while dropping its SQuAD F1 from 64.56 to 51.27, and DepGraph pruning raises CPU latency in every controlled comparison. The model family, compression level, and recovery method must therefore be selected according to the downstream task profile rather than by applying a uniform compression policy. Tables~\ref{tab:deployment_guidance_llm} and~\ref{tab:deployment_guidance_image} summarize task-specific deployment recommendations derived from our empirical results and can serve as a guideline for selecting compressed models under different edge constraints.

\subsection{Stage-wise Model Re-selection}
\label{subsec:stage_wise_model_reselection}

Our deployment recommendation from the image results is that, for the six segmentation candidates, model selection has to be repeated after every deployment stage rather than fixed at the pre-deployment baseline. The model that gives the best full-size task performance does not remain the best after compression, and the model that gives the best latency is not the best choice under thermal constraints. We therefore treat edge deployment as a stage-wise filtering process: first select strong full-size candidates, then re-rank them after pruning or quantization, and finally re-rank them again after real-device deployment.

\subsubsection{Evidence}

Table~\ref{tab:1} shows that FPN--EfficientNet-B3 is the strongest full-size image model before compression, achieving the highest mIoU of 0.6728 and the highest mean accuracy of 75.59\%. However, after structured pruning at $S=0.5$, this model no longer remains the best compressed candidate, as its mIoU drops to 0.5252 and its accuracy drops to 61.22\%. In contrast, DeepLabV3+--ResNet-50 becomes the strongest pruned model, retaining 0.6258 mIoU and 70.97\% accuracy while reducing its size from 106.71 MB to 30.30 MB. This shows that pre-deployment accuracy alone is not sufficient for selecting an edge model. Table~\ref{tab:post-dp} further shows that deployment ranking changes again when latency and device-level behavior are considered. For example, DeepLabV3+--EfficientNet-B3 achieves the fastest wall latency at pruning ratio 0.9 with 11.5 s, while U-Net--EfficientNet-B3 reaches the lowest CPU latency of 5.9 s and the lowest reported temperature of 49.1$^\circ$C at the same pruning ratio. Thus, some models are better latency candidates, whereas others may be more suitable under thermal constraints. These segmentation results support stage-wise, multi-objective selection over choosing by baseline accuracy alone.

\section{Conclusion and Future Work}
\label{sec:conclusion_future_work}

This article connects model compression to practical edge deployment for Artificial Intelligence of Things (AIoT) systems. First, we reviewed recent deployment-oriented studies and extracted practical guidelines from prior work on pruning, quantization, tensor decomposition, and knowledge distillation. Then, following these guidelines, we conducted an empirical study across multiple downstream tasks, including SQuAD, BoolQ, Natural Questions, and image segmentation, using GPU, CPU, and Raspberry Pi platforms. Our results show that compression effectiveness is strongly task, model and hardware-dependent: a method that shrinks the stored checkpoint can still increase deployed latency, and the model that performs best before compression does not remain the best choice after pruning, quantization, recovery, or real-device deployment. Finally, we provided deployment recommendations that map downstream tasks to use cases, constraints, and empirical evidence.
Several future directions remain open.

\begin{itemize}
    \item First, on-device LLM inference is often motivated by acceleration techniques such as KV-cache optimization, speculative decoding, prompt caching, token pruning, early exiting, and memory-aware scheduling. However, these techniques are rarely selected using a unified prefill--decode-level analysis. A useful future direction is to develop an optimization framework that profiles the prefill and decode stages separately, estimates the latency--memory--accuracy trade-off of different acceleration schemes, and recommends the most suitable configuration for a given resource-constrained device. Such a framework would be particularly useful for Raspberry Pi-class and mobile AIoT platforms, where memory bandwidth and decoding latency dominate practical LLM inference.

    \item Second, our image-model experiments show that a stock PyTorch FX-graph-mode post-training quantization workflow can fail to produce a compressed artifact for complex image segmentation architectures, and it does not report which operators were actually converted. A model-agnostic quantization framework that automatically traces complex architectures, identifies quantizable and non-quantizable regions, reports unsupported operators, and applies safe quantization without breaking the computation graph would be a valuable contribution to the edge-AI community. Such a framework should not only quantize supported layers, but also provide an explicit operator-level deployment report so that practitioners can understand why compression succeeds or fails for a given architecture.

    \item Finally, extending this methodology to additional domains would test how far these recommendations carry. In this work, we focused on compact LLMs and image models; however, AIoT systems increasingly include automatic speech recognition, audio event detection, multimodal sensing, time-series forecasting, medical signal analysis, and robotics workloads. Extending the same deployment-aware methodology to ASR and other domain-specific models would reveal new latency, memory, accuracy, and energy trade-offs.
\end{itemize}

\FloatBarrier
\bibliographystyle{ACM-Reference-Format}
\bibliography{Ref}
\end{document}